\documentclass[utf8]{frontiersSCNS} % for Science, Engineering and Humanities and Social Sciences articles
\usepackage{url,hyperref,lineno,microtype,subcaption}
\usepackage[onehalfspacing]{setspace}

\usepackage[normalem]{ulem}

\usepackage{multirow}
\def\keyFont{\fontsize{8}{11}\helveticabold }
\def\firstAuthorLast{Mohseni {et~al.}} %use et al only if is more than 1 author
\def\Authors{Mahdi Mohseni\,$^{1}$, Volker Gast\,$^{2}$ and Christoph Redies\,$^{1,*}$}
\def\Address{$^{1}$Experimental Aesthetics Group, Institute of Anatomy I, Jena University Hospital, University of Jena, School of Medicine, Jena, Germany \\
$^{2}$Department of English and American Studies, University of Jena, Jena, Germany  }
\def\corrAuthor{Dr. Christoph Redies\\
Institute of Anatomy I\\
Jena University Hospital\\
D-07740 Jena\\
Germany\\
Phone: +49 - 3641 - 9396 100}

\def\corrEmail{christoph.redies@med.uni-jena.de}

\newcommand\Tstrut{\rule{0pt}{2.6ex}}       % "top" strut
\newcommand\Bstrut{\rule[-0.9ex]{0pt}{0pt}} % "bottom" strut
\usepackage{xcolor}

\let\footnoterule\relax

\begin{document}
\onecolumn
\firstpage{1}

\title[Analysis of Global Structure in Literary and Non-Literary Texts]{Comparative Computational Analysis of Global Structure in Canonical, Non-Canonical and Non-Literary Texts
}

\author[\firstAuthorLast ]{\Authors} %This field will be automatically populated
\address{} %This field will be automatically populated
\correspondance{} %This field will be automatically populated

\extraAuth{}% If there are more than 1 corresponding author, comment this line and uncomment the next one.
%\extraAuth{corresponding Author2 \\ Laboratory X2, Institute X2, Department X2, Organization X2, Street X2, City X2 , State XX2 (only USA, Canada and Australia), Zip Code2, X2 Country X2, email2@uni2.edu}

\maketitle

\begin{abstract}

\noindent This study investigates global properties of literary and non-literary texts.
Within the literary texts, a distinction is made between canonical and non-canonical works.
The central hypothesis of the study is that the three text types (non-literary, literary/canonical and literary/non-canonical) exhibit systematic differences with respect to structural design features
as correlates of aesthetic responses in readers.
To investigate these differences, we compiled a corpus containing texts of the three categories of interest, the Jena Textual Aesthetics Corpus.
Two aspects of global structure are investigated, variability and self-similar (fractal) patterns, which reflect long-range correlations along texts.
We use four types of basic observations,
(i) the frequency of POS-tags per sentence,
(ii) sentence length,
(iii) lexical diversity in chunks of text, and
(iv) the distribution of topic probabilities in chunks of texts.
These basic observations are grouped into two more general categories, 
(a) the low-level properties (i) and (ii), which are observed at the level of the sentence (reflecting linguistic decoding), and
(b) the high-level properties (iii) and (iv), which are observed at the textual level (reflecting comprehension).
The basic observations are transformed into time series,
and these time series are subject to multifractal detrended fluctuation analysis (MFDFA), giving rise to three statistics:
(i) the degree of fractality,
(ii) the fractal dimension (width of the fractal spectrum), and
(iii) the degree of asymmetry of the fractal spectrum.
Our results show that  low-level properties of texts are better discriminators than high-level properties, for the three text types under analysis. Canonical literary texts differ from non-canonical ones primarily in terms of variability. Fractality seems to be a universal feature of text, more pronounced in non-literary than in literary texts. While some of our results are hard to interpret from a literary point of view, we surmise that our findings reflect an important design feature of text, the distribution of discourse modes (Narrative, Report, Description, Information, Argument).
Beyond the specific results of the study, we intend to open up new perspectives on the 
 experimental study of textual aesthetics.

%%% 

\tiny
 \keyFont{ \section{Keywords:} 
 fractality, self-similarity, multifractal DFA, variability, POS tagging, sentence length, lexical diversity, topic modeling} %All article types: you may provide up to 8 keywords; at least 5 are mandatory.
\end{abstract}

%%%%%%%%%%%%%%%%%%%%%%%%%%%%%%%%%%%%%%%%%%%%%%%%%%%%%%%%%%%%%%%%%%%%%%%%%%
%%%%%%%%%%%%%%%%%%%%%%%%%%%%%%%%%%%%%%%%%%%%%%%%%%%%%%%%%%%%%%%%%%%%%%%%%%
\section{Introduction}\label{sec:introduction}

The goal of the present work is to objectively measure differences in global structure between texts of three categories (non-literary, literary/canonical and literary/non-canonical). To this aim, we introduce and validate various statistical measures that describe the self-similarity (fractality) and variance of specific semantic properties across individual texts. By comparing literary with non-literary texts, and within the group of literary texts, canonical and non-canonical ones, we provide a basis for understanding aesthetic responses of human readers to global properties of text. This line of research follows a similar approach that has been applied successfully in visual aesthetics during the last decade \citep{brachmann-redies2017}. By applying objective statistical measures to literary prose, we introduce computational text analysis into the field of experimental aesthetics \citep{Chatterjee2014,Jacobs2015}. 
%Our present work thus extends previous investigations in this direction of research. 

The founder of experimental aesthetics, Gustav Theodor Fechner, proposed that the aesthetic appeal of visual objects is based on stimulus properties that can be measured in an objective (formalistic) way \citep{fechner1876}. A few decades later, Clive Bell (\citeyear{Bell1914}) speculated that visual artworks possess a ‘significant form’, which has the potential to elicit an aesthetic response in beholders across art periods and cultures. 
This notion has been opposed by some (post-)modern philosophers, art critics and psychologists \citep[for example, see][]{Danto1981,leder2004}. They advanced conceptual theories which stipulate that cultural context and content are crucial and sufficient to evaluate artworks. In this modern view of aesthetic experience, traditional concepts like ‘beauty‘ no longer play a prominent role. Pushing this view to the extreme, it has been claimed that any physical object can be a work of art, as long as experts declare it to be an artwork in the appropriate cultural context \citep{Danto1981}. Nevertheless, the idea that beautiful artworks possess an intrinsic formal structure keeps reappearing even in modern aesthetic theories \citep[for visual stimuli, see][]{Arnheim1974}. Contemporary versions of such formalist theories \citep{taylor2011, redies2007} postulate that large sets of visual artworks share image properties that reflect a specific physical structure. For example, it has been suggested that traditional artworks may share regularities in the layout of basic pictorial elements, such as luminance gradients and their orientations \citep{taylor2011,redies2007}. These and other structural image properties have been measured in visual artworks in recent years, and some of them can be used to distinguish traditional artworks from non-art images \citep[for reviews, see][]{graham2010,redies2015}. 

In visual aesthetics, a particular focus has been on global image properties of artworks. In contrast to local image properties, such as luminance contrast or color at a given location in an image, global image properties reflect summary statistics of pictorial elements or their relations to each other across an image \citep[for a review, see][]{brachmann-redies2017}. Global statistical image properties seem particularly suitable for studying aesthetic properties because aesthetic concepts such as ‘balanced composition‘ \citep{mcmanus1985}, ‘good Gestalt‘ (Arnheim, 1974), or ‘visual rightness‘ \citep{locher1999} all refer to global image structure \citep{redies2017}. Examples of global properties characteristic of artworks are an intermediate degree of complexity \citep{berlyne1974,forsythe2011}, specific color features \citep{palmer2010,mallon2014}, a fractal-like image structure \citep{taylor2011}, statistical regularities in the Fourier domain \citep{graham2007,redies2007}, luminance statistics \citep{graham2008}, curvature \citep{bar2006,bertamini2016} and regularities in edge orientation distribution \citep{redies2012,redies2017}. Moreover, traditional visual artworks exhibit a high richness and high variability of low-level features that are computed by a Convolutional Neural Network \citep[CNN;][]{brachmann2017}.

Even though there is no long-standing tradition in empirical textual aesthetics comparable to that of visual aesthetics, the question of objective, measurable properties of texts that reflect aesthetic perception has been raised in various contexts, more or less explicitly. The assumption
that aesthetic appeal can be measured is most obvious for poetry, with its interplay
of meaning and form as manifested in rhythm and rhyme, and other aspects of poetic form, e.g. alliteration \citep[cf. for instance][]{jakobson1960,Leech1969,Jacobs2015,Jacobs2016,Vaughan2016,Konig2017,Menninghaus2017,Egan2019}. Relevant studies of
prose texts refer to more abstract (global) structural properties of the texts. 
For example, global statistical properties such as complexity and entropy have been used to study the regularity \citep{mehri2016,Hernandez2017} and the quality of texts \citep{Febres2017}. Fractal analysis has been applied to literary texts \citep{Drozdz2015,mehri2016,chatzigeorgiou2017}, and fractal patterns have been observed in both Western \citep{Drozdz2016} and Chinese literature \citep{Yang2016,Chen2018}.
\citet[796]{Cordeiro2015} claim that “there is a fractal beauty in the text produced by humans” and “that its quality is directly proportional to the degree of self-similarity.”

While the results obtained in the aforementioned studies are still tentative, they suggest that text has structural correlates of aesthetic experience in reading.
The starting point of the present study is the hypothesis that these correlates are comparable to those found in vision, and we focus on two global properties, i.e. variability and fractality. Our hypothesis of an analogy between visual and linguistic processing is based on the assumption widely made in cognitive linguistics that ``linguistic structure is shaped by domain-general processes'' \citep[23]{diessel2019}
such as figure-ground segregation and processes of memory retrieval.
In other words, linguistic processing is based on the same type of brain activity as the processing of other types of sensory input.
The analogy has obvious limitations though.
Image data are three-dimensional – two-dimensional matrices with the luminance/color signals as the third dimension – whereas textual data are \textit{prima facie} one-dimensional when regarded as strings of characters (though even silent reading implies prosody,
adding a second dimension, cf. \citealt{Grossetal2014}).

Related to this, the processing of propositional information is incremental
\citep{Verhuizen2019},
with new 
information constantly being added while earlier information fades out,
being summarized and generalized in the process.
A part of the aesthetic experience is thus less immediate
and relates to higher levels of processing. Still, reading implies low-level processing
activity which can be expected to trigger certain responses to the input signal in the brain.

While the processing of visual information is rather well understood, there is little experimental evidence about
how information is processed during reading. The `classic' model -- the LaBerge/Samuels model of automatic information processing in reading \citep[cf.][]{LaBergeSamuels1974,Samuels1994} --
assumes four components, (i) visual memory (VM), (ii) phonological memory (PM), (iii) semantic memory (SM)  and (iv) episodic memory (EM).
VM and PM are closely connected to sensory experience, i.e. visual and acoustic perception, and they are the input gates to processing in reading.
Semantic memory is not only the place where
``individual word meanings are produced", but also ``where the comprehension of written messages occurs"
\citep[710]{Samuels1994}. It is thus also responsible for the linguistic process of decoding, including
the processing of morphology (word structure) and syntax (sentence structure). Episodic memory -- or explicit memory, as we call it -- 
is the place where propositional information is stored, and it is ``responsible for putting a time, place and context tag on events and knowledge" \citep[710]{Samuels1994}.

We assume two levels of processing in reading, a low level of linguistic decoding, and a high level of integrating the propositional information conveyed in the input signal into explicit memory,
i.e. comprehension. This is largely analogous to models of language processing
for spoken language (see for instance \citealt{BornkesselSchlesewsky2006, martin2020}).
Reading (as well as the processing of spoken language) obviously implies bottom-up as well as top-down processes, and the continuous integration of linguistic information and world knowledge \citep{Verhuizen2019}.
For example, the propositional content of a message is a function of its components, while the interpretation of any given word is heavily context-dependent and thus influenced by the surrounding information. This is particularly obvious for figurative language, cf. for instance I.A. Richards' theory of metaphor \citep{Richards1936}. While low-level and high-level processing interact in the reading experience, we assume that they are cognitively distinct and  have different neural substrates. Evidence for this assumption can be found in experimental work, e.g. using eye-tracking methodology \citep{Weissetal2018,CookWei2019}.

We hypothesize that the three text types under analysis differ in terms of aesthetic
experience during reading.
Literary texts are intended to evoke an aesthetic response while non-literary texts are primarily informative. Moreover, we assume that the long-term ``success" of canonical literature reflects, to some extent, perceptual or cognitive processes in the reading experience, though literary success obviously
depends on other factors as well (cf. \citealt{UnderwoodSellers2016}).

Given the time-distributed nature of information processing in reading, aesthetic experience is hard to measure experimentally \citep[see, for instance,][for discussion]{CookWei2019}.
We therefore pursue an observational, rather than experimental approach, assuming that
aesthetic responses to a text have structural correlates in the text itself.
For a systematic quantitative analysis we have compiled a corpus of literary and non-literary texts,
the Jena Textual Aesthetics Corpus (cf. Section \ref{sec:corpus}). The literary texts of this corpus
are classified into canonical and non-canonical ones.
We use the Corpus of the Canon of Western Literature \citep{Green2017}, which was compiled on the basis of \cite{Bloom1994} (\textit{The Western Canon: The Book and School of the Ages}), as a benchmark for canonicity,\footnote{Obviously, a literary canon reflects not only properties of the texts themselves, but also attitudes held by the compilers, and aesthetic attitudes to literary works are highly culture specific and, to some extent, learned. These potential objections notwithstanding, we hypothesize that canonical literature is distinguished from non-canonical literature with respect to certain (measurable global) properties reflecting preferences of non-professional as well as  professional readers, such as critics and literary scholars. It is an interesting question, beyond the scope of this study, whether a different canon -- e.g., a canon of African American Literature \citep[cf.][]{GatesMcKay2004} -- would yield different results.} and use information
from international Wikipedia Websites as additional evidence for the higher prestige of
canonical (as opposed to non-canonical) authors.
In comparing literary texts and non-literary texts, we assume that literary texts are aesthetically more pleasing than non-literary text. The framing of information and linguistic structures can be expected to be different between these text types. Literary and non-literary texts differ in terms of the distribution of `modes of discourse' \citep{smith2003} or, for brevity's sake, `discourse modes' (cf. also the more traditional term `rhetorical mode'; see \citealt{Newman1827}). Non-literary texts seem to cover all of the modes distinguished by \citet{smith2003}, i.e.
the temporal modes Information and Argument as well as the temporal modes 
Narrative, Report and Description.
Literary texts mostly consist of narrative and descriptive parts but also contain elements of internal communication (dialogue, monologue, thoughts). Note also that the differentiation of the modes can be expected to be clearer in non-literary
texts, which often have a schematic structure, reflected in labeled (sub-)sections. We surmise that such differences between literary and non-literary texts have reflexes in global structural properties of the texts.

The remainder of this article is organized as follows. Firstly, we provide a list of measurable text properties that may contribute to aesthetic experience in reading (Section \ref{sec:text-representation}). Based on these properties, the texts are transformed into time series. Secondly, we introduce methods that capture the distribution of these properties across the time series, with a particular focus on two features (variability and fractality/self-similarity), which we consider as potential mediators of aesthetic experience in reading (Section \ref{sec:methods}). Thirdly, to examine whether any of these properties are associated with the aesthetics of reading, we compare canonical with non-canonical literary texts as well as literary texts with non-literary ones. 
% {\color{red}By doing so, we assume that the three text categories differ in their aesthetic claim.} 
The three sub-corpora analyzed are introduced in Section \ref{sec:corpus}. Fourthly, in Section \ref{sec:analysis}, we study how well our new analytical methods and text features can distinguish between the three text categories. Note that this first pilot study is restricted to a subset of text properties and analysis methods that seemed particularly promising to us. A complete analysis of all combinations of properties and methods is beyond the scope of the present work. Finally, in Section \ref{sec:discussion}, we discuss the implications of our preliminary findings and outline research questions that can be addressed with the proposed methods in the future.

%%%%%%%%%%%%%%%%%%%%%%%%%%%%%%%%%%%%%%%%%%%%%%%%%%%%%%%%%%%%%%%%%%%%%%%%%%
%%%%%%%%%%%%%%%%%%%%%%%%%%%%%%%%%%%%%%%%%%%%%%%%%%%%%%%%%%%%%%%%%%%%%%%%%%
%\section{Converting prose text into a numerical format}\label{sec:text-representation}
\section{Measurable properties of text}\label{sec:text-representation}

The central hypothesis of this study is that the aesthetic appeal of texts correlates with measurable structure of the texts. Such properties can be derived from various types of measurements. While all the measurements that we used for our analysis represent global properties of the texts, the basic units of observations are located at different levels of processing. As mentioned in Section \ref{sec:introduction}, we distinguish two levels of processing. The lower level of processing concerns the task of linguistic decoding, which is largely automatic and resorts to implicit knowledge. Aesthetic experience at this level is connected to lexical meaning (e.g. is the imagery congruent and appealing?) and grammatical structure (e.g. is a sentence easy to process?). The higher level of processing concerns the integration of propositional information into explicit memory (comprehension).

In this section, we point out some measurable properties of text that we assume to trigger responses in the human reader: two types of low-level properties (frequencies of part-of-speech tags and sentence length; Subsections \ref{sec:text-representation:pos} and \ref{sec:text-representation:sentlength}); and 
two types of high-level properties (lexical diversity and topic distribution; Subsections \ref{sec:text-representation:lexical} and \ref{sec:text-representation:topic}). These are the properties that were used in our exploratory studies (cf. Section \ref{sec:analysis}). In addition, we point out two types of properties that can be applied at various levels of text, i.e. embedding vectors and language models (Subsections \ref{sec:word-embeddings} and  \ref{sec:language-model}). We will not report any results obtained for the latter  properties but we suggest that they may be used in future studies. We represent each property as time series that can then be subjected to an analysis of their global structural features, such as variances and fractal features. 

%%%%%%%%%%%%%%%%%%%%%%%%%%%%%%%%%%%%%%%%%%%%%%%%%%%%%%%%%%%%%%%%%%%%%%%%%%

\subsection{Part-of-Speech Tags}\label{sec:text-representation:pos}

Part-of-speech tags, commonly abbreviated as `POS-tags', represent the syntactic class of a word. To some extent, they reflect syntactic structure. At the most general level, POS-tags classify words into major classes such as `noun', `verb', `adjective' etc., but depending on the specific tagset used, more fine-grained distinctions can be made (e.g. between singular and plural nouns). For the present study, we used the Stanford Tagger (version 3.6.0)\footnote{https://nlp.stanford.edu/static/software/tagger.shtml}, which assigns words to the classes distinguished by the Penn Treebank tagset.\footnote{https://www.ling.upenn.edu/courses/Fall\_2003/ling001/penn\_treebank\_pos.html} 
We determined the frequencies of specific POS-tags per sentence, giving rise to (sets of) time series.

%%%%%%%%%%%%%%%%%%%%%%%%%%%%%%%%%%%%%%%%%%%%%%%%%%%%%%%%%%%%%%%%%%%%%%%%%%
\subsection{Sentence Length}\label{sec:sentence-length}\label{sec:text-representation:sentlength}

Sentence length has been used as a measurement for the study of fractality before by \citet{Drozdz2016}, though not for a comparison of text types. It is easy to extract and provides some structural information at the sentence level. Sentence length is highly correlated with the frequency of specific POS-tags in sentences, e.g. the (absolute) number of nouns. In our corpus (Section \ref{sec:corpus}), the average Spearman correlation coefficient of these two time series is 0.86. However, as will be seen, the variability and fractal analysis of these two text properties performs differently in classification of the three different text categories of our corpus (see Section \ref{sec:analysis}).

%%%%%%%%%%%%%%%%%%%%%%%%%%%%%%%%%%%%%%%%%%%%%%%%%%%%%%%%%%%%%%%%%%%%%%%%%%

\subsection{Lexical Diversity}\label{sec:text-representation:lexical}

The choice of words is one of the most perspicuous properties of a text, and a 
rich vocabulary is often regarded as a hallmark of good authorship. For example,
\citet{simonton1990} claims that lexical diversity  correlates with ``aesthetic success".
He analyzed Shakespeare's sonnets and showed that there is a vocabulary shift from the more ``obscure" to the more popular sonnets.
Vocabulary and the richness of lexicon has also been found useful in the assessment of writers' proficiency, for instance in research on second language acquisition \citep[see][]{laufer1995,zareva2005,yu2009}.
 Several metrics have been proposed for measuring lexical diversity.
Type-Token Ratio (TTR) is the simplest one, in which the number of distinct words (types) is divided by the length of the text. However, TTR is highly affected by text length. In our experiments (cf. Section \ref{sec:analysis}), we use the Measure for Textual Lexical Diversity \citep[MTLD;][]{McCarthy2010}, which is more robust because it is less sensitive to text length.

%%%%%%%%%%%%%%%%%%%%%%%%%%%%%%%%%%%%%%%%%%%%%%%%%%%%%%%%%%%%%%%%%%%%%%%%%%
\subsection{Topic Distribution}\label{sec:text-representation:topic}

Topic modeling is a method used to analyze the content of texts by revealing hidden topics of documents in a collection. We are interested in the changes of topic distribution along a text (rather than  the global topics of a text). To extract the distribution of topics from a text, the text is split into segments and then, to infer the topic distribution, a topic modeling method is applied. Latent Dirichlet Allocation (LDA) \citep{blei2003,griffiths2004}, the most widely used topic modeling method, or an extension of it can be used for this purpose. For long-range correlations and variability analysis (see Section \ref{sec:methods}) one can convert the topic distribution of the text into a time series by computing a distance measure, e.g. the Jensen–Shannon divergence, of topic representations of adjacent chunks. It is also possible to analyze the topic distribution matrix in terms of its variability. In Section \ref{sec:analysis}, we will show that patterns of topic distribution vary across texts and can thus be informative for recognizing categories of texts.

%%%%%%%%%%%%%%%%%%%%%%%%%%%%%%%%%%%%%%%%%%%%%%%%%%%%%%%%%%%%%%%%%%%%%%%%%%
\subsection{Language Model}\label{sec:language-model}

Language modeling is an essential part of many language processing tasks such as machine translation, summarization and speech recognition. A language model computes the probability of a sequence of words and predicts the probability of the next word \citep{jurafsky2009}. 
Language models capture both semantic and structural information, as the probability
for a given word to occur is a function of both the surrounding structure and the semantic
context. A time series can be created, for example, by calculating the probabilities of consecutive text segments, such as sentences or paragraphs.

%%%%%%%%%%%%%%%%%%%%%%%%%%%%%%%%%%%%%%%%%%%%%%%%%%%%%%%%%%%%%%%%%%%%%%%%%%
\subsection{Embedding vectors}\label{sec:word-embeddings}

More precise ways of making the distribution of linguistic segments measurable have been provided by recent
advances in automatic language processing. By applying neural models, distributed 
representations of words and text have been developed,
resulting in an improvement of almost all natural language processing tasks.
Embedding vectors -- n-dimensional vectors of floats -- represent
the distribution of a linguistic segment and allow for the computation of (dis)similarities between segments.
A wide variety of models have been proposed to represent text at the level of sub-word, word, sentence, etc. \cite[for example, see][]{pennington2014,bojanowski2017,devlin2018}.
For a study of global text properties, word embeddings can be converted to time series using distance measures,
e.g. cosine distance, and analyzed by fractal analysis methods or processed directly using neural or non-neural algorithms. 

\vspace{12pt}
%%%%%%%%%%%%%%%%%%%%%%%%%%%%%%%%%%%%%%%%%%%%%%%%%%%%%%%%%%%%%%%%%%%%%%%%%%
%%%%%%%%%%%%%%%%%%%%%%%%%%%%%%%%%%%%%%%%%%%%%%%%%%%%%%%%%%%%%%%%%%%%%%%%%%
\section{Global measures of variability and self-similarity}\label{sec:methods}

In the present section, we introduce ways of analyzing the time series of text properties that were proposed in the previous section. We focus on two global statistical features (variability and self-similarity). These properties have previously been used in visual aesthetics and have been associated with artworks and other visually pleasing stimuli (see Section \ref{sec:introduction}). 

Variability reflects the degree to which a particular feature (e.g., edge orientation or color) is likely to vary across an image. 
Self-similarity is closely related to fractality and scale-invariance. This property reflects the degree to which parts of an image have features similar to the image as a whole, i.e., an image is self-similar if it shows similar features at different scales of resolution. 
To analyze variability and fractality
several methods are available and some of them will be described in the following subsections. Where they have been used in text analysis before, we will briefly outline their previous usage. In Section \ref{sec:analysis}, we will then apply these measures to analyze the time series that were introduced in Section \ref{sec:text-representation} in our corpus of texts (Section \ref{sec:corpus}).

Global statistical measures have been applied to texts before. For example, linguistic laws such as Zipf's and Heaps' laws were proposed to provide insights into the internal structure of text \citep[for example, see][]{Baayen2002,Serrano2009}.
Zipf's law establishes a power law distribution between word frequencies and ranks of words (according to their frequencies) in texts. It states that a small number of word types accounts for a high percentage of word tokens in a text, while the number of low-frequent words is very high. Another empirical law, Heaps' law, assumes a power law distribution between the vocabulary size, i.e. the number of distinct words, and the number of words in a document or a corpus. Heaps' law states that the ratio of the vocabulary size to the length of document(s) decreases drastically as more text is added. These global features, however, have not been used in the context of text aesthetic. These linguistic laws ignore relations between text components and are supposed to be universally valid for different genres of texts.

%%%%%%%%%%%%%%%%%%%%%%%%%%%%%%%%%%%%%%%%%%%%%%%%%%%%%%%%%%%%%%%%%%%%%%%%%%
\subsection{Variance}
The variability of a property can be measured simply by computing its variance. The variance of a random variable $X$ is
\begin{equation}
    \mathcal{V}(X) = E[(X-\mu)^2]
\end{equation}
$E[.]$ denotes the expected value and $\mu$ is the population mean. The variance of, for example, the distribution of sentence length reflects the amount of variation in the length of sentences across a text. Despite its mathematical simplicity, we will see that variance performs effectively in the classification of text categories (Section \ref{sec:analysis}). 

%%%%%%%%%%%%%%%%%%%%%%%%%%%%%%%%%%%%%%%%%%%%%%%%%%%%%%%%%%%%%%%%%%%%%%%%%%
\subsection{Entropy-Based Methods} 

Entropy, which is related to variability, measures uncertainty or (ir)regularity of a state or phenomenon represented by a random variable. If $X$ is a discrete random variable with a set of possible values $\{x_1, x_2, \cdots, x_n\}$ and a corresponding probability function $P(X)=\{P(x_i), P(x_2), \cdots, P(x_n)\}$, the entropy of $X$ is defined as:
\begin{equation*}
    H(X) = -\sum_{i=1}^{n}P(x_i) \log_b P(x_i)
\end{equation*}
Entropy is zero when the state is certain and it is highest when the all possibilities are equally likely to occur, i.e. when uncertainty is maximal. The basic formula of entropy or its extensions have been utilized for text analysis previously.

\cite{Rosso2009} applied statistical complexity and entropy quantifiers to a collection of poems and plays. Their analyses revealed that poems have a higher complexity than plays and Shakepeare's work is interestingly more homogeneous than that of his contemporaries and is exceptionally close to the average use of words in that time period.
\cite{Chang2017} defined the information-based energy, combined from the relative temperature and information Shannon entropy, to quantify text complexity and an author's performance. Applying this method to texts of an English and an Chinese author, Shakespeare and Jin Yong, they showed that their more popular works have higher information-based energy. 
\cite{Hernandez2017} used an entropy-based method, called approximate entropy, to measure the degree of irregularity or randomness in a time series. They applied this method to 14 different languages which belong to four linguistic families: Romance, Germanic, Slavic and Uralic. They showed that the languages exhibit different levels of irregularity which were similar for languages that belonged to the same family.
The entropy of word distributions can also be informative for comparing
different types of languages in term of word ordering. \cite{montemurro2016} used entropy-based measures to show that word ordering is highly similar over several language families. \cite{Febres2017} studied entropy and symbolic diversity of literary texts of Nobel and non-Nobel laureates in English and Spanish. While they presented some results to show that there is a correlation between these global statistical properties and the quality of writing, they did not classify different groups of texts. 

%%%%%%%%%%%%%%%%%%%%%%%%%%%%%%%%%%%%%%%%%%%%%%%%%%%%%%%%%%%%%%%%%%%%%%%%%%
\subsection{Box Counting}

There are several methods to measure fractality and the scaling behavior of structures. These methods typically represent measurements at different scales. Fractal analysis techniques have been widely applied to images \citep{wendt2007,Li2009,wendt2009,Ji2013}, including artworks \citep{taylor2002,redies2007,Spehar2016}. They are therefore of special interest of analyzing aesthetic phenomena (cf. Section \ref{sec:introduction}).

One of the most widely used fractal analysis methods is box counting, which is mathematically straightforward and easy to apply. Given an object $S$, for a $\delta>0$ the smallest possible number of subsets with a diameter of at most $\delta$, $N_\delta (S)$, which covers $S$, is found. For 1d objects, subsets are rulers and $\delta$ is their length. For 2d objects, subsets are boxes and $\delta$ is their area, and so forth. The growth ratio of $N_\delta (S)$, as $\delta \rightarrow 0$, reflects the degree of fractality of $S$. If $N_\delta (S)$ can be approximated by 
\begin{equation*}
    N_\delta (S)\simeq c\delta^{D_B}
\end{equation*}
for a constant $c$, then $D_B$ is called the box-counting dimension and shows how complex $S$ is. 

\cite{mehri2016} applied this method to seven famous text books and computed their degree of fractality by averaging the fractality degrees of word occurrences. The results revealed that all texts are fractal and their fractal dimensions differed slightly. Fractality patterns of time series
sometimes do not lend themselves to analysis with
a single scaling measure.
If different subsets of a time series exhibit different types of scaling behavior, the time series is multifractal.
\cite{chatzigeorgiou2017} used box counting to find the origin of multifractality in the word-length representation of texts in several Western languages. They showed that the long-range correlations in natural language are related to the clustering feature of long words, i.e. rare and often highly informative content words.

\vspace{6pt}
%%%%%%%%%%%%%%%%%%%%%%%%%%%%%%%%%%%%%%%%%%%%%%%%%%%%%%%%%%%%%%%%%%%%%%%%%%
\subsection{Wavelet-Based Methods}

Fractal analysis methods based on wavelets are another family of techniques for studying scale-invariant properties of signals \citep{muzy1993, wendt2007,leonarduzzi2016}. The wavelet transform (WT) is a method to analyze non-stationary signals. The WT of a signal $X$ is defined as \citep{mallat1999}:
\begin{equation*}
    T_\psi [X](a, t_0) = \frac{1}{a}\int_{-\infty}^{+\infty}X(t)\psi(\frac{t-t_0}{a})dt,
\end{equation*}
and it describes the content of $X$ around a time parameter $t_0$ and a scale parameter $a$. $\psi$ is the analyzing wavelet whose $n+1$ first moments are zero, i.e. $\int_\mathbb{R}t^n\psi(t)dt=0$, which makes the WT insensitive to possible polynomial trends  of order $n$ in the signal, something which is necessary for multifractal analysis \citep{muzy1994,Arneodo1995}. The WT modulus maxima (WTMM) is a well-known method for analyzing multifractality and it is based on the WT coefficients. WTMM is defined by the local maxima $\mathcal{L}(a)$ of $|T_\psi [X](a, t)|$ according to a given scale $a$. Then the following partition function is defined:
\begin{equation*}
    Z(q, a)= \sum_{l\in \mathcal{L}(a)} |T_\psi [X](a, t)|^q \sim a^{\tau(q)}
\end{equation*}
If the signal is monofractal, $\tau(q)$ is independent of $q$. For multifractal signals, the scaling behavior cannot be explained with one value, so, $\tau(q)$ changes for different values of $q$. Based on WT and WTMM, other methods have been extended for discrete and multi-dimensional time series \citep[for example, see ][]{wendt2007,leonarduzzi2016}. Although wavelet-based methods have been applied to a variety of fields, they have been rarely used in text processing. \cite{leonarduzzi2017} applied the wavelet p-leader method to the sentence-length series of novels that were written either for young people or adults. The authors showed that the latter category is more diverse in terms of its degree of multifractality.

%%%%%%%%%%%%%%%%%%%%%%%%%%%%%%%%%%%%%%%%%%%%%%%%%%%%%%%%%%%%%%%%%%%%%%%%%%
\subsection{Detrended Fluctuation Analysis}\label{sec:methods:mfdfa}

Detrended Fluctuation Analysis (DFA) \citep{Peng1994} and its extension Multi-Fractal DFA (MFDFA) \citep{kantelhardt2002} have been widely used in studying long-range correlations in a broad range of research fields, such as biology \citep{Das2016}, economics \citep{Caraiani2012}, music \citep{Sanyal2016} and animal song \citep{roeske2018}. 
MFDFA can be related to Fourier spectral analysis and both methods provide similar results for the degree of fractality \citep{Heneghan2000}. 
Moreover, MFDFA has a theoretical and practical connection with wavelet-based methods \citep{leonarduzzi2016}. 

MFDFA is a straightforward, efficient and numerically stable method for multifractal analysis \citep{Oswiecimka2006}. In the present work, we will apply this method to the fractal analysis of texts. Given a time series $X = x_1, x_2, \cdots, x_N$, MFDFA can be summarized as follows:
\begin{enumerate}
    \item Subtract the mean and compute the cumulative sum, called the profile, of the time series:
    
    $Y(i) = \sum_{k=1}^{i} [x_k - <x>], i=1, \cdots, N$
    \item Divide the profile of the signal into $N_s=N/s$ windows for different values of $s$
    \item Compute the local trend, $Y'$, which the best fitting line (or polynomial), in each window
    \item Calculate the mean square fluctuation of the detrended profile in each window $v$, $v=1,\cdots,N_s$ :
    
    $F^2(s, v)=\frac{1}{s}\sum_{i=1}^{s}[Y(s \times (v-1) + i)-Y'(s \times (v-1) + i)]^2$

    \item Calculate the $q$th order of the mean square fluctuation:
    
    $F_q(s) = \{\frac{1}{N_s}\sum_{v=1}^{N_s}[F^2(s, v)]^{q/2}\}^{1/q}$
    \item Determine the scaling behavior of $F_q(s)$ versus $s$: $F_q(s)\sim s^{h(q)}$
\end{enumerate}

The procedure is 
equivalent
to DFA if $q$ is fixed at 2. For monofractal time series, $h(q)$ is independent of $q$. If a time series is stationary, $h(2)$ is equal to the Hurst Exponent, a well-known measure in fractal analysis studies. We refer to this value as $\mathcal{H}$, which is the fractal degree of the time series. In the remainder of this text, wherever we use `Hurst exponent' we refer this value, even though the time series may not be stationary. For uncorrelated time series, in which each event is independent of other events, $\mathcal{H}\simeq 0.5$. With $\mathcal{H}$ increasing above 0.5, the time series is more fractal. In the opposite direction, if $\mathcal{H}<0.5$ the time series is called anti-persistent in which a large value in the time series is most likely followed by a small value, and vice versa. 

\begin{figure}
    \centering
    \includegraphics[scale=0.44]{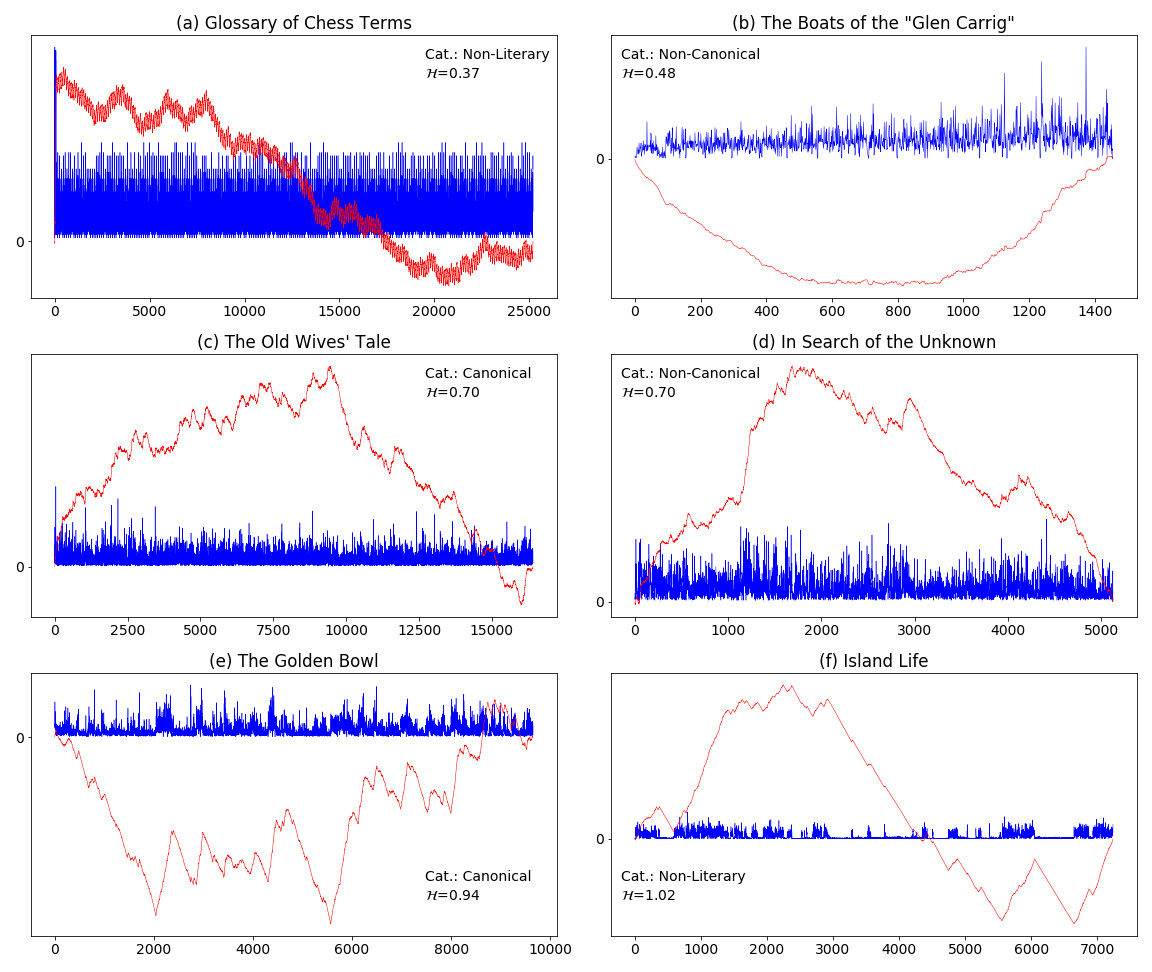}
    \caption{Sentence length time series (blue) and their profiles (red; cumulative sum of mean centered series) of some example texts in the corpus. The time series have been scaled up 20 times to show more detail. The category (Cat.) and the fractal degree, $\mathcal{H}$, of each text is shown inside each panel.  (a) Glossary of Chess Terms by Gregory Zorzos with the lowest fractal degree in our corpus. (b) Boats of the ``Glen Carrig" by William Hope Hodgson with $\mathcal{H}=0.48$, a non-canonical text with the lowest value among the literary books. (c) Women in Love by D. H. Lawrence, the median of canonical texts. (d) In Search of the Unknown by  Robert W. Chambers, representing the median of non-canonical texts. (e) The Golden Bowl by Henry James with the highest fractal degree among canonical texts. (f) Island Life by  Alfred Russel Wallace, a non-literary text, with the highest fractal degree in the whole corpus.}
    \label{fig:h-example}
\end{figure}

To explain $\mathcal{H}$ better, we show the sentence-length time series of a few cases in our corpus (Section \ref{sec:corpus}) as well as the profile of each time series in Figure \ref{fig:h-example} (see step 1 of MFDFA in above). Figure \ref{fig:h-example}(a) represents the time series of the  Glossary of Chess Terms by Gregory Zorzos, which is one of the texts in the non-literary categories of our corpus. This is not a usual text but a dictionary-like book consisting of a list of term-values. This example is an anti-persistent text with $\mathcal{H}=0.37$  and an extreme case in the corpus with the lowest fractal degree. Figure \ref{fig:h-example}(b) belongs to The Boats of the ‘‘Glen Carrig’’ by William Hope Hodgson. With $\mathcal{H}=0.48$, this book has the second lowest $\mathcal{H}$ and the closest one to 0.5, which shows that there is almost no correlation among the elements of its time series. This book is categorised as a non-canonical text in our corpus. As a side note, the lower bound of fractality for sentence-length time series of canonical texts in the corpus starts from $\mathcal{H}=0.58$, which is the value measured for Old Mortality by Walter Scott. In Figure \ref{fig:h-example}(c) and \ref{fig:h-example}(d) we show the plots of one canonical and one non-canonical literary book with a median degree of fractality, within the relevant category/sub-corpus. For both The Old Wives' Tale by Arnold Bennett, a canonical text, and In Search of the Unknown by  Robert W. Chambers, a non-canonical text,  $\mathcal{H}=0.70$. Figure \ref{fig:h-example}(e) represents the time series of a canonical text with the highest fractal degree ($\mathcal{H}=0.94$) in the corpus, namely The Golden Bowl by Henry James. Finally, the text with the highest fractality value in the entire corpus is Island Life by Alfred Russel Wallace, which is a non-literary book with $\mathcal{H}=1.02$.

From $h(q)$, one can compute the fractal dimension and the fractal asymmetry, two metrics that represent the fractal complexity of the time series. From $h(q)$, the Hölder exponents $\alpha=h(q)+qh'(q)$ and the singularity spectrum $f(\alpha)=q[\alpha-h(q)]+1$ are computed. Then, the fractal dimension is defined as $\mathcal{D}=\alpha_{max}-\alpha_{min}$ (cf. \citealt{kantelhardt2002,Drozdz2016}). $\alpha_{min}$ and $\alpha_{max}$ denote the beginning and the end of $f(\alpha)$, respectively. The fractal asymmetry is also determined from $f(\alpha)$:
\begin{equation*}
    \mathcal{A} = \frac{\Delta\alpha_L-\Delta\alpha_R}{\Delta\alpha}
\end{equation*}
where $\Delta\alpha_L=\alpha_0-\alpha_{min}$ and $\Delta\alpha_R=\alpha_{max}-\alpha_{0}$ \citep{Drozdz2015}. $\alpha_0$, corresponding to $q=0$, usually points to the peak of the $f(\alpha)$ curve. It is also obvious that $\mathcal{D}=\Delta\alpha_L + \Delta\alpha_R$. In Section \ref{sec:analysis}, we will use the three values (fractal degree [$\mathcal{H}$], fractal dimension [$\mathcal{D}$] and fractal asymmetry [$\mathcal{A}$]) as a basis to classify the three categories of text (canonical, non-canonical and non-literary).

\begin{figure}
    \centering
    \includegraphics[scale=0.45]{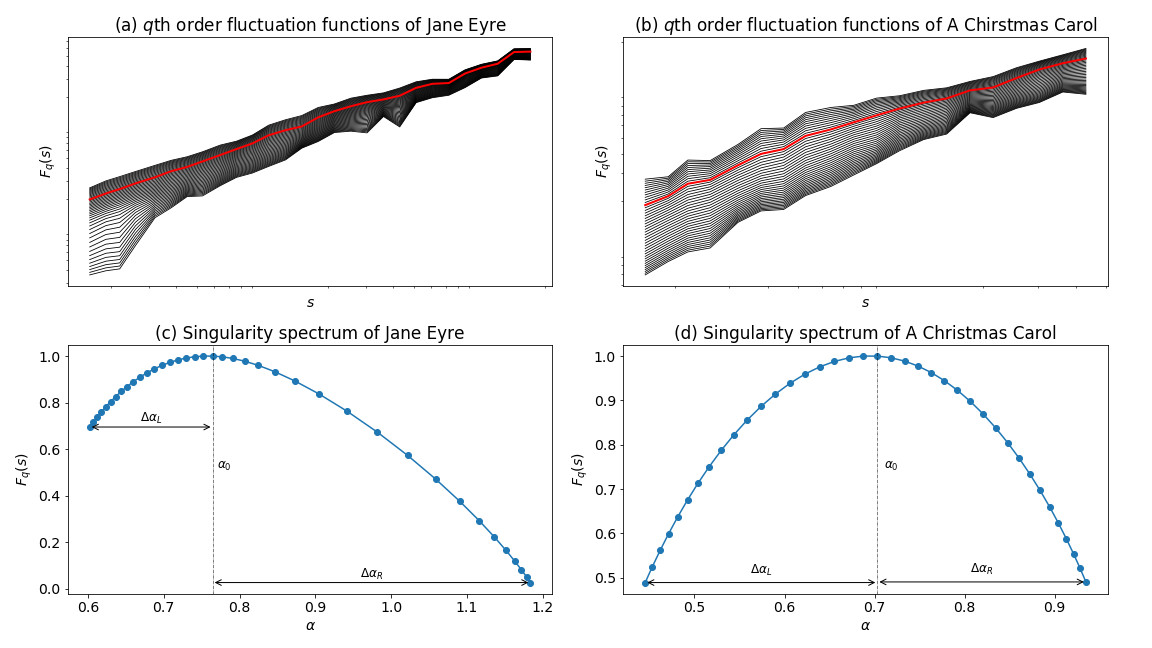}
    \caption{$q$th order of mean square fluctuation of sentence-length time series of (a) Jane Eyre written by Charlotte Bronte and (b) A Christmas Carol written by Charles Dickens. The plots show $F_q(s)$ for different scales of $s$ and for different values of $q$ ranging from $-5$ to $5$ with step 0.25. $\mathcal{H}$ is the slope of best fitting line to the red curve corresponding to $q$ of value 2, which is equal to the output of the DFA method. Singularity spectra of the two texts are plotted in panels (c) and (d).
    $\alpha_0$ indicates the peak of each curve. The width of the curve, $\Delta\alpha = \Delta\alpha_L + \Delta\alpha_R$, known as the fractal dimension ($\mathcal{D}$), shows how multifractal a times series is. Here, both texts are highly multifractal. The fractal asymmetry ($\mathcal{A}$) of the curve is calculated from $\Delta\alpha_L$ and $\Delta\alpha_R$. It is asymmetrical for Jane Eyre, but symmetrical for A Christmas Carol. See the text for more details.}    \label{fig:asymmetry-example}
\end{figure}

To illustrate these concepts visually, we show the results of the fractal analysis for canonical texts by Charlotte Bront\"e and Charles Dickens in Figure \ref{fig:asymmetry-example}. The two texts have been converted to time series by using the sentence-length property (see Section \ref{sec:sentence-length} for details). Fig.\ref{fig:asymmetry-example}(a) and Fig.\ref{fig:asymmetry-example}(b) show $F_q(s)$ for different values of $q$ ranging from $-5$ to $5$. 
It is obvious that the slope of the lines, $h(q)$, changes as $q$ changes. This result indicates that the texts are multifractal. Jane Eyre written by Charlotte Bront\"e has a fractal dimension $\mathcal{D}=0.58$ indicating a high degree of multifractality. The figure also shows that the time series has a high fractal asymmetry, $\mathcal{A}=0.48$ (Fig. \ref{fig:asymmetry-example}(c)). Figure \ref{fig:asymmetry-example}(d) presents the singularity spectrum for A Christmas Carol by Charles Dickens with $\mathcal{D}=0.49$ and $\mathcal{A}=-0.02$, which indicate that the time series of the text is multifractal and (almost) symmetrical.

DFA has been used for text analysis previously, in particular for sentence-length analysis. For example, \cite{Drozdz2015} applied MFDFA to sentence-length time series in comparison with other natural time series (e.g., the discharge of the Missouri river and sunspot number variability) and non-natural time series (e.g., stock market and Forex index prices). The results suggest that natural languages possess a multifractal structure that is comparable to other natural and non-natural phenomena. \cite{Yang2016} investigated long-range correlations in sentence-length series in a famous classic Chinese novel, based on the number of characters in each sentence. This study showed that there was a long-range correlation, though it was weak. A diachronic fractality analysis of word-length in Chinese texts spanning 2,000 years revealed two different long-range correlations regimes for short and large scales \citep{Chen2018}. An analysis of the fractal dimension of sentence-length time series in several Western literary texts revealed that, although most literary texts show a long-range correlation, the dimension of fractality can be quite different among them, ranging from monofractal to highly multifractal structure \citep{Drozdz2016}. Although sentence length can be measured in various ways, e.g., by the number of characters or words in unlemmatized and lemmatized texts, the different ways yield robust results that have comparable distributions and similar patterns of long-range correlations \citep{Vieira2018}.

%%%%%%%%%%%%%%%%%%%%%%%%%%%%%%%%%%%%%%%%%%%%%%%%%%%%%%%%%%%%%%%%%%%%%%%%%%

\subsection{Fractality and Cross-Correlation Analysis}

Fractal analysis can be extended to analyzing more than one time series, in order to find relations between fractal behaviors of multiple time series. Detrended Cross-Correlation Analysis (DCCA) \citep{podobnik2008} and Multi-Fractal Detrended Cross-Correlation Analysis (MFDCCA) \citep{Jiang2011} are two methods for analyzing correlations between two time series. \cite{ghosh2019} applied MFDCCA, also known as MFDXA, to study correlations between two Tagore’s poems, one written in Bengali and one in English. They found a nonlinear correlation between the poems. In a similar study, birdsong and human speech were compared by computing the mutual information decay of signals and it was concluded that the two vocal communication signals have similar dynamics \citep{Sainburg2019}.

%%%%%%%%%%%%%%%%%%%%%%%%%%%%%%%%%%%%%%%%%%%%%%%%%%%%%%%%%%%%%%%%%%%%%%%%%%
\subsection*{}

In our experiments, as mentioned above, we focus on the variability and self-similar (fractal) patterns of text properties. We will use variance to analyze variability of texts in our corpus. To analyze fractal patterns, we will focus on the most widely used method (MFDFA). The result of these analyses is used for classification of the categories of text that are introduced in the following section.

%%%%%%%%%%%%%%%%%%%%%%%%%%%%%%%%%%%%%%%%%%%%%%%%%%%%%%%%%%%%%%%%%%%%%%%%%%
%%%%%%%%%%%%%%%%%%%%%%%%%%%%%%%%%%%%%%%%%%%%%%%%%%%%%%%%%%%%%%%%%%%%%%%%%%
\section{The corpus}\label{sec:corpus}

As mentioned in Section \ref{sec:introduction}, we use three sub-corpora representing three major categories: a corpus of canonical literary texts, a corpus of non-canonical literary texts, and a corpus of non-literary texts.

The canonical literary sub-corpus comprises 77 English prose texts, written by 29 different authors, from Period C (1832--1900) and Period D (20th century) of the Corpus of Canonical Western Literature (\citet{Green2017}; cf. also Section \ref{sec:introduction}). We selected those texts from the corpus that were sufficiently long for our analysis (at least 35K words).

The non-canonical literary texts were downloaded from e-book publishing sites in the internet. We primarily used www.smashwords.com, an e-book distributor website that is catering to classic texts, independent authors and small press. It offers a large selection of books from several genres and allows downloads in various formats. The books are classified into `Fiction', `Nonfiction', `Essays', `Poetry' and `Screenplays'. We selected random books from various prose genres, using the site's filter to make sure that the books had a specified minimal length as of canonical texts. 
We further supplemented the corpus of non-canonical books with the lowest rated books on www.goodreads.com and www.feedbooks.com, as well as books with the lowest rates of downloads on the Project Gutenberg site. These books are in the public domain, written mostly between 1880 and 1930 and more than 45K words in length. In this way, we obtained 95 books of non-canonical literature (from as many authors in each case). 
We made sure to collect non-canonical texts from the same time period as for our canonical sub-corpus to minimize the effect of phenomena such as language change on our analyses. However, collecting ``low-quality'' non-canonical texts from one century back is not easy as these texts have probably not been preserved or, at least, not digitized. If they survived and are still read, they are likely to be of relatively high quality. Therefore, our non-canonical sub-corpus can be regarded as a topnotch non-canonical and, thus, close to the canonical sub-corpus, which makes our analysis more difficult. Nevertheless, the non-canonical texts selected by us are clearly non-canonical in the sense that they currently do not belong to any canon of literature such as the one that we used for the selection of canonical texts. 

\begin{figure}
    \centering
    \includegraphics[scale=.6]{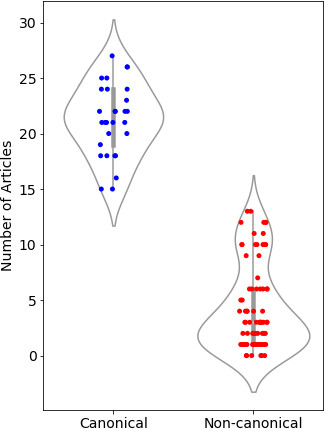}
    \caption{Number of articles in the top 30 language editions of Wikipedia for authors in the canonical (blue) and non-canonical (red) sub-corpora.}
    \label{fig:wiki-authors}
\end{figure}

As another discriminating factors between canonical and non-canonical texts, we counted the number of articles that each author has in the top 30 language editions of Wikipedia. This measure is evidence for the international reputation of an author. Figure \ref{fig:wiki-authors} shows a strip plot for all authors in each category. There is a clear separation between the authors of the two groups. All authors of canonical texts have at least 15 articles each in the 30 Wikipedia editions. In the non-canonical category, each author has up to 13 articles at the most; for the majority of authors, the number is less than 5.
These numbers provide independent evidence for the higher degree of prestige \citep{UnderwoodSellers2016} of canonical authors, in comparison to non-canonical authors.

To construct our non-literary category we relied on Project Gutenberg. We downloaded all non-literary books and randomly selected 133 books from different genres such as architecture, astronomy, geology, geography, philosophy, psychology, sociology. To increase the diversity, we added the first two volumes of The Encyclopedia Britannica by University of Cambridge and a text called Glossary of Chess Terms by Gregory Zorzos. The latter text was added to our corpus because of its strange and interesting fractal behavior, as discussed in the previous section and shown in Figure \ref{fig:h-example}. The texts of the two literary categories, with the exception of the last one, were published during similar time periods.
    
\begin{figure}
    \centering
    \includegraphics[width=.7\textwidth]{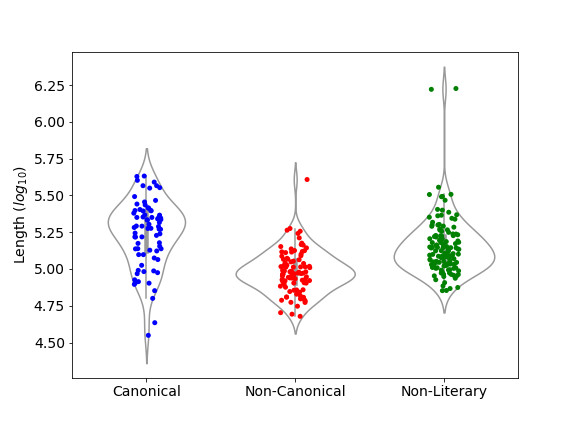}
    \caption{Strip plot of book length (number of tokens, $log_{10}$) for the three categories of text in our corpus (canonical [blue], non-canonical [red] and non-literary [green].)}
    \label{fig:corpus-statistics}
\end{figure}

Figure \ref{fig:corpus-statistics} shows the number of tokens per book (i.e. the length of the texts) on a logarithmic scale, grouped into the three categories of interest. The three categories differ somewhat in the distribution of text lengths. The two very long non-literary texts shown in the figure are the two volumes of The Encyclopedia Britannica. It is important to realize that the exact length of a text does not affect the results of our experiments, given that the texts are sufficiently long to be analyzed robustly for their variability or fractal properties.

The texts were tagged manually to eliminate material not belonging to the core text, such as tables of contents and indices. Headers were left in the text, as they are potentially informative. Moreover, the texts were semi-automatically cleaned up using regular expressions to identify (and re-join) hyphenated words at the end of a line. 
Information about the entire corpus, which we named the `Jena Textual Aesthetics Corpus', is provided in Supplementary Table S1. 

The core hypothesis of the present study is that the three different text categories under analysis -- non-literary texts, literary/canonical and literary/non-canonical ones -- differ in terms
of aesthetic responses in the reader, and that these aesthetic responses have measurable correlates
in global text structure (cf. Section \ref{sec:text-representation}).
The Jena Textual Aesthetics Corpus allows us to test this hypothesis, as it contains
samples of text from the three categories of interest. 
In order to compare the text categories, we carried out two binary classification tasks.
The first task (Task 1) is to separate the literary from the non-literary works. The second task (Task 2) consists in separating the canonical literary texts from the non-canonical ones. The results are reported in Section \ref{sec:analysis}.

%%%%%%%%%%%%%%%%%%%%%%%%%%%%%%%%%%%%%%%%%%%%%%%%%%%%%%%%%%%%%%%%%%%%%%%%%%
%%%%%%%%%%%%%%%%%%%%%%%%%%%%%%%%%%%%%%%%%%%%%%%%%%%%%%%%%%%%%%%%%%%%%%%%%%
\section{Analysis and Classification Results}\label{sec:analysis}

In this section, we present an analysis of the variability and fractality of structural properties of text as well as the classification results for the three text categories in our corpus. For reasons of space, we have not calculated all measures or features that were proposed in Sections \ref{sec:text-representation} and \ref{sec:methods}. Instead, we present the results for four textual properties only (POS-tag frequencies, sentence length, lexical diversity and topic distributions). The first two properties are low-level properties while the latter two are high-level properties (see Section \ref{sec:text-representation}). The time series derived from these properties have been subjected to a variance analysis as well as an analysis of their fractality, which has been restricted to MFDFA, the most widely-used fractal analysis methods of time series. The original results presented in the present work thus serve as a proof of concept and do not provide a complete coverage of all possible types of analyzing variability and fractality. To the best of our knowledge, the present work is the first to analyze fractality of text using lexical diversity, the frequency of POS-tags and topic distributions, and to utilize variability and fractality analysis to classify text categories.

\subsection{Converting Texts into Time Series}

POS-tags, sentence length, lexical diversity and topic distribution were introduced in Section \ref{sec:text-representation}. For the sake of reproducibility of our results, we will provide further details in the present section on how we calculated the relevant measures and thereby converted the texts to time series.

To convert a text into a time series of POS-tag frequencies, we determined the number of each specific tag in the sentences of the text (see Subsection \ref{sec:text-representation:pos}). In our analysis, we focused on nouns, adjectives, verbs, and pronouns. Other POSs either did not yield interesting results or occurred too infrequently. For the annotations we used the Stanford POS-tagger \citep{toutanova2003}. For the calculations, we included all types of nouns, i.e. singular as well as plural nouns and proper names. Several types of verb forms -- for example, base forms, past tense forms, gerund, past participles -- were all treated as verbs. Adjective includes simple, comparative as well as superlative adjectives. Pronouns are either personal or possessive. We thus obtained four different time series derived from the frequency of POS-tags.

Sentence length is another property of sentences that can be used to generate time series for the purpose of fractal analysis (see Subsection \ref{sec:text-representation:sentlength}). To determine the length of a sentence, we first used the NLTK-package \citep{nltk} to sentence-tokenize the texts. The length of each sentence is the number of its tokens. Punctuation marks were not removed, and were counted as elements of sentences.

Lexical diversity measures the richness of vocabulary of a text (see Subsection \ref{sec:text-representation:lexical}). To convert a text into a time series of lexical diversity values, we first segmented the text into chunks that were 100 tokens long, which seems like a good compromise between reliability of the calculations, and the required minimal length for fractal analysis. We then computed MTLD values \citep{McCarthy2010} for each chunk to obtain a time series for this feature.

Topic modeling is a high-level analysis of text that focuses on the content conveyed (see Subsection \ref{sec:text-representation:topic}). To extract the topic distribution of a text, we first segmented the text into coherent chunks using the TopicTiling algorithm \citep{riedl2012}. Then, we applied the LDA algorithm \citep{blei2003,griffiths2004} to all chunks of all texts in the corpus, thus obtaining a topic model. The number of topics, one of the hyperparameters of LDA, was set to 100. The resulting topic model is a statistical model of 100 topic that shows the importance of each word in a topic. Afterwards, the topic model was applied to each chunk of a text to infer the distribution of the 100 topics (the `topic probabilities'). In order to convert the vector of topic probabilities to a time series, we calculated the Jensen–Shannon divergence of the topic representations of adjacent chunks.

\subsection{Analysis of Variance and Fractality}\label{sect:varfract}

After generating the time series for the seven text properties for all texts, we calculated the variance, $\mathcal{V}$, as a measure of how variable each text property was across each text. Moreover, we used MFDFA to calculate the following fractal features for each text: the degree of fractality ($\mathcal{H}$), the fractal dimension ($\mathcal{D}$) and the degree of fractal asymmetry ($\mathcal{A}$) (see Subsection \ref{sec:methods:mfdfa}). 
As Kolmogorov-Smirnov tests revealed that some of the data were not normally distributed, the data was entered into a Wilcoxon test to assess the differences between the three original subcorpora, supplemented by non-parametric Mann-Whitney tests for all (post-hoc) pairwise comparisons.
The median values of the variances and fractal features are shown in Table \ref{table:confidence-interval} for all three subcorpora of text (canonical, non-canonical and non-literary). In addition, we  obtained the same variables for both types of literary text (canonical and non-canonical texts) together, as we distinguish two classification tasks: the distinction between literary versus non-literary texts (Task 1), and between canonical versus non-canonical texts (Task 2; see Section \ref{sec:corpus}).

\newcommand{\daggerone}{{1}}
\newcommand{\daggertwo}{{2}}
\newcommand{\daggerthree}{{3}}

\begin{table}
    \footnotesize
    \setlength\tabcolsep{1.0pt}
    \centering
    \caption{Median values and 95\% confidence intervals (in parentheses) for all combinations of text properties (columns: lexical diversity [MTLD], POS-tags [noun, verb, adjective, and pronoun], sentence length, topic distribution) and features (rows: variance [$\mathcal{V}$], degree of fractality [$\mathcal{H}$], fractal dimension [$\mathcal{D}$] and fractal asymmetry [$\mathcal{A}$]). Each feature is analyzed for two tasks: Literary (Lit.; $N$ = 172) vs. non-literary (Non-Lit.; $N$ = 135) texts (Task 1), and canonical (Can.; $N$ = 77) vs. non-canonical (Non-Can.; $N$ = 95) texts (Task 2). The asterisks indicate whether the differences between the two text categories of a given task are statistically significant (Mann-Whitney test; *, $p \leq 0.05$; **, $p \leq 0.01$; and ***, $p \leq 0.001$). In addition, canonical and non-canonical texts are compared separately with non-literary texts; the superscript numbers show whether the differences are significant (Mann-Whitney test; $^\daggerone$, $p \leq 0.05$; $^\daggertwo$, $p \leq 0.01$; and $^\daggerthree$, $p \leq 0.001$). }
    \begin{tabular}{cc|ccccccc}
        \multicolumn{2}{c}{} & Noun & Verb & Adjective & Pronoun & Sentence Length & MTLD & Topic Distribution \Tstrut\\\hline
        % \multirow{6}{0.5cm}{$\sqrt{\mathcal{V}}$} & Lit. & 3.34 (3.16, 3.58) & 2.54 (2.37, 2.69) & 1.54 (1.48, 1.66) & 1.81 (1.73, 1.91) & 14.9 (13.6, 16.6) & 19.4 (19.0, 19.8) & 0.067 (0.066, 0.068) \\
        % & Non-Lit. & 4.32 (4.08, 4.50) & 2.75 (2.66, 2.89) & 2.04 (1.98, 2.13) & 1.36 (1.20, 1.45) & 17.5 (17.0, 18.3) & 17.9 (17.2, 18.7) & 0.069 (0.068, 0.073) \\
        % & & *** & & *** & *** & *** & *** & ***
        % \\\cline{2-9}
        \multirow{6}{0.3cm}{$\mathcal{V}$} & Lit. & 11 (10, 13) & 6.5 (5.6, 7.2) & 2.3 (2.1, 2.8) & 3.3 (3.0, 3.6) & 220 (184, 277) & 376 (361, 391) & 4.5e-3 (4.4e-3, 4.6e-3) \\
        & Non-Lit. & 19 (17, 20) & 7.5 (7.0, 8.3) & 4.2 (3.9, 4.5) & 1.9 (1.4, 2.1) & 305 (290, 336) & 322 (295, 348) & 4.8e-3 (4.6e-3, 5.3e-3) \\
        & & *** & ** & *** & *** & *** & *** & ***
        \\\cline{2-9}
        % & Can. & 3.87 (3.69, 4.13) & 3.01 (2.69, 3.15) & 1.81 (1.72, 1.96) & 2.06 (1.91, 2.24) & 17.9 (17.2, 19.2) & 19.8 (19.4, 20.2) & 0.069 (0.067, 0.070) \\
        % & Non-Can. & 3.02 (2.84, 3.16) & 2.24 (2.12, 2.45) & 1.37 (1.29, 1.44) & 1.64 (1.56, 1.73) & 12.8 (12.1, 13.9) & 18.9 (18.6, 19.5) & 0.065 (0.063, 0.067) \\
        % & & *** & *** & *** & *** & *** &  & *** \\\hline
        & Can. & 15 (14, 17)$^\daggertwo$ & 9.0 (7.2, 9.9) & 3.3 (3.0, 3.9)$^\daggertwo$ & 4.3 (3.7, 5.0)$^\daggerthree$ & 321 (296, 367) & 390 (375, 408)$^\daggerthree$ & 4.8e-3 (4.5e-3, 4.9e-3) \Tstrut\\
        & Non-Can. & 9.1 (8.1, 10)$^\daggerthree$ & 5.0 (4.5, 6.0)$^\daggerthree$ & 1.9 (1.7, 2.1)$^\daggerthree$ & 2.7 (2.4, 3.0)$^\daggerthree$ & 163 (145, 194)$^\daggerthree$ & 357 (345, 381)$^\daggertwo$ & 4.2e-3 (4.0e-3, 4.5e-3)$^\daggerthree$ \\
        & & *** & *** & *** & *** & *** & ** & *** \\\hline
        
        \multirow{6}{0.3cm}{$\mathcal{H}$} & Lit. & 0.714 (0.706, 0.725) & 0.66 (0.65, 0.67) & 0.685 (0.677, 0.695) & 0.67 (0.66, 0.68) & 0.70 (0.69, 0.71) & 0.65 (0.64, 0.66) & 0.63 (0.62, 0.65) \\
        & Non-Lit. & 0.69 (0.67, 0.71) & 0.72 (0.70, 0.76) & 0.72 (0.70, 0.74) & 0.71 (0.70, 0.73) & 0.73 (0.70, 0.75) & 0.68 (0.66, 0.70) & 0.66 (0.61, 0.69) \\
        & & * & *** & *** & *** &  & *** & 
        \\\cline{2-9}
        & Can. & 0.72 (0.70, 0.73) & 0.67 (0.65, 0.68)$^\daggerthree$ & 0.69 (0.68, 0.70)$^\daggerone$ & 0.67 (0.65, 0.68)$^\daggerthree$ & 0.70 (0.68, 0.71) & 0.64 (0.63, 0.66)$^\daggerone$ & 0.64 (0.61, 0.66) \Tstrut\\
        & Non-Can. & 0.71 (0.70, 0.73) & 0.66 (0.64, 0.67)$^\daggerthree$ & 0.68 (0.67, 0.69)$^\daggertwo$ & 0.67 (0.65, 0.68)$^\daggerthree$ & 0.70 (0.68, 0.71) & 0.65 (0.64, 0.66)$^\daggertwo$ & 0.63 (0.61, 0.65) \\
        & & & & & & & & \\\hline
        
        \multirow{6}{0.3cm}{$\mathcal{D}$} & Lit. & 0.30 (0.26, 0.32) & 0.20 (0.17, 0.21) & 0.31 (0.28, 0.33) & 0.67 (0.66, 0.68) & 0.26 (0.24, 0.28) & 0.20 (0.17, 0.22) & 0.19 (0.18, 0.21) \\
        & Non-Lit. & 0.26 (0.23, 0.31) & 0.37 (0.34, 0.40) & 0.30 (0.25, 0.37) & 0.71 (0.70, 0.73) & 0.34 (0.29, 0.42) & 0.20 (0.18, 0.22) & 0.25 (0.20, 0.28) \\
        & &  & *** &  & *** & *** &  &  
        \\\cline{2-9}
        & Can. & 0.34 (0.32, 0.36) & 0.23 (0.19, 0.25)$^\daggerthree$ & 0.32 (0.28, 0.34) & 0.67 (0.65, 0.68)$^\daggerthree$ & 0.31 (0.26, 0.35) & 0.20 (0.16, 0.22) & 0.18 (0.15, 0.21)$^\daggerone$ \Tstrut\\
        & Non-Can. & 0.26 (0.23, 0.30) & 0.16 (0.15, 0.20)$^\daggerthree$ & 0.29 (0.27, 0.33) & 0.67 (0.65, 0.68)$^\daggerthree$ & 0.23 (0.20, 0.25)$^\daggerthree$ & 0.21 (0.17, 0.23) & 0.20 (0.18, 0.21) \\
        & & *** & *** &  &  & *** &  &  \\\hline

        \multirow{6}{0.3cm}{$\mathcal{A}$} & Lit. & 0.03 (-0.05, 0.09) & 0.09 (0.02, 0.14) & 0.04 (-0.03, 0.07) & 0.09 (-0.01, 0.17) & 0.08 (0.01, 0.14) & 0.11 (0.06, 0.18) & 0.15 (0.06, 0.23) \\
        & Non-Lit. & 0.24 (0.12, 0.41) & 0.61 (0.55, 0.68) & 0.55 (0.48, 0.66) & 0.36 (0.28, 0.48) & 0.56 (0.45, 0.69) & 0.15 (0.04, 0.28) & 0.09 (-0.04, 0.27) \\
        & & *** & *** & *** & *** & *** &  &  
        \\\cline{2-9}
        & Can. & 0.09 (0.00, 0.13)$^\daggerone$ & 0.10 (0.04, 0.16)$^\daggerthree$ & 0.04 (-0.03, 0.08)$^\daggerthree$ & 0.16 (0.05, 0.21)$^\daggertwo$ & 0.13 (0.04, 0.23)$^\daggerthree$ & 0.10 (0.03, 0.20) & 0.20 (-0.02, 0.27) \Tstrut\\
        & Non-Can. & -0.04 (-0.13, 0.06)$^\daggerthree$ & 0.07 (-0.03, 0.20)$^\daggerthree$ & 0.04 (-0.07, 0.11)$^\daggerthree$ & -0.01 (-0.07, 0.20)$^\daggerthree$ & 0.02 (-0.02, 0.12)$^\daggerthree$ & 0.12 (-0.06, 0.24) & 0.15 (0.05, 0.26) \\
        & &  &  &  &  &  &  &  \\\hline
        
    \end{tabular}
    \label{table:confidence-interval}
\end{table}

Table \ref{table:confidence-interval} shows that none of the text properties (POS-tag frequencies, sentence length, lexical diversity and topic probabilities) results in significantly different median values for all features (variance and fractality measures) in both tasks. The high-level properties (MTLD and topic distributions) do not vary significantly across text types for the fractal features. However, the variance ($\mathcal{V}$) is significantly different for all features in both tasks. 
Strikingly, $\mathcal{V}$ values are always higher for non-literary texts than for literary texts, except for the values obtained from frequencies of pronouns, and MTLD values. This difference is mainly driven by non-canonical texts. $\mathcal{V}$ values for canonical texts range in between those for non-literary and non-canonical texts. In some cases (verb frequencies, sentence length and topic distributions), the values for canonical texts are not significantly different from those of non-literary texts, but higher than the values for non-canonical texts.

In summary, in terms of $\mathcal{V}$, canonical texts are more similar to non-literary texts than to non-canonical texts. Only for the frequency distribution of pronouns and MTLD values do the canonical texts exhibit the highest values, followed by non-canonical texts and, with even lower values, by non-literary texts. Note that the magnitude of the variances does not reflect the magnitude of mean values for the text properties (cf. Suppl. Table S2 for the mean values).

The degree of fractality, $\mathcal{H}$, is of similar magnitude (closer to 0.5) for all text properties for canonical and non-canonical literary texts. By contrast, the $\mathcal{H}$ values for non-literary texts are generally higher than for either type of literary text (canonical or non-canonical), with the exception of the frequencies of nouns, sentence length and topic distributions. These results suggest that a lower degree of long-range correlations might be a uniform characteristic of literary texts as opposed to non-literary texts, regardless of the status of the literary texts as canonical or non-canonical. 

The values for the fractal dimension, $\mathcal{D}$, are significantly higher for the frequencies of verbs and pronouns as well as sentence length in non-literary as opposed to literary texts. A comparison of canonical and non-canonical literary texts reveals that the $\mathcal{D}$ values of canonical texts are consistently higher than or equal to the values for non-canonical texts, even though this tendency reaches statistical significance only for the frequencies of nouns and verbs, as well as sentence length.

The degree of asymmetry, $\mathcal{A}$, does not differ between canonical and non-canonical texts. For low-level properties, literary texts are rather symmetrical (i.e. close to 0), and $\mathcal{A}$ is higher for non-literary texts than for literary texts. For the higher-level properties (MTLD values and topic distributions), $\mathcal{A}$ values do not vary across the three sub-corpora.

To summarize the observations made above, canonical texts show more variability with respect to the properties measured in our study than non-canonical texts, and are, in this respect, more similar to non-literary texts. However, the lower degree of fractality ($\mathcal{H}$) suggests that the two types of literary texts display a lower degree of long-range correlations than non-literary texts do. Moreover, canonical texts tend to be more multifractal than non-canonical texts in terms of the frequencies of nouns and verbs, as well as for sentence length (higher $\mathcal{D}$). Unlike in the case of non-literary texts, the fractal spectra of literary texts  are rather symmetrical ($\mathcal{A}$ is closer to 0). 

\begin{figure}[t]
     \centering
     \includegraphics[scale=0.43]{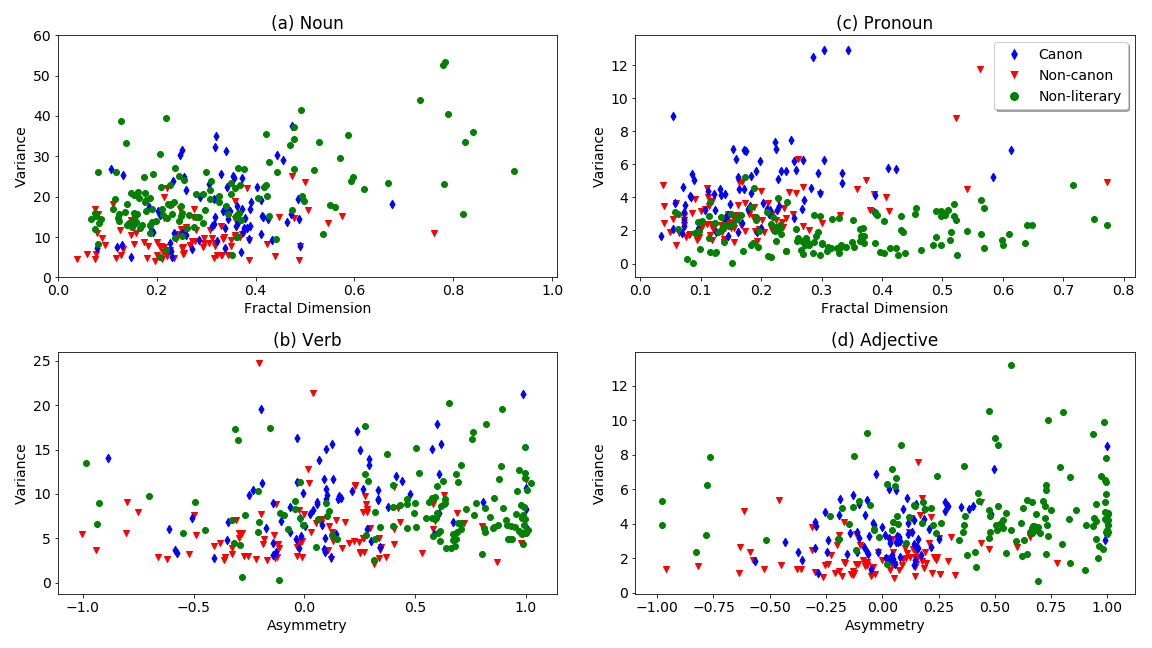}
    %  \caption{H and variance of noun time series.}
    \caption{Scatter plots of  variance (y-axis) and fractal features of POS-tags (x-axis). (a) Degree of fractality ($\mathcal{H}$) and the variance of noun time series. (b) Fractal dimension ($\mathcal{D}$) and the variance of verb time series. (c) Fractal dimension ($\mathcal{D}$) and the variance of pronoun time series. (d) Fractal asymmetry ($\mathcal{A}$) and the variance of adjective time series. Each dot represents one text from our corpus. For color coding of the text categories, see insert in (b).}
     \label{fig-pos}
\end{figure}

The individual values for the variance (y-axis) and fractal features (x-axis) for selected text properties are shown as scatter plots in Figure \ref{fig-pos} to illustrate the separation and overlap between the different text categories. For this figure, we chose plots that showed a relatively clear separation of the text categories by subjective visual inspection. Figure \ref{fig-pos}(a) shows the fractal dimension and the variance of noun times series. As stated above (Table \ref{table:confidence-interval}), the variances for non-canonical texts tend to be lower than those of the other two categories. Fig. \ref{fig-pos}(b) depicts the fractal dimension and the variance of pronoun frequencies; it shows that literary texts tend to have a higher variance compared to non-literary texts. Both Figure \ref{fig-pos}(a) and Figure \ref{fig-pos}(b) confirm that non-literary texts scatter in a wider range of the fractal dimension. In Figure \ref{fig-pos}(c) and Figure \ref{fig-pos}(d), the variances of verb and adjective time series are plotted as a function of the degree of asymmetry. Fractal patterns of non-literary texts are more asymmetrical (higher $\mathcal{A}$). Again, canonical texts exhibit a wider scatter, as variance is higher compared to non-canonical texts, which suggests more diverse usage of language structures in canonical texts. The behavior of non-literary texts varies across the tags. For example, the texts scatter more widely in the plot of adjectives (Fig. \ref{fig-pos}[d]), while their pronoun variances cover a narrower range (Fig. \ref{fig-pos}[d]), since pronouns are not so frequent in non-literary texts (Suppl. Table S2). Figure \ref{fig-pos} also illustrates that non-literary texts have a more complex fractal pattern and spread more broadly along the fractal feature (x-)axes. Non-literary texts tend to show a higher fractal degree and more fractal asymmetry than literary texts.

%%%%%%%%%%%%%%%%%%%%%%%%%%%%%%%%%%%%%%%%%%%%%%%%%%%%%%%%%%%%%%%%%%%

\subsection{Classification}\label{sec:classification}
While a statistical analysis of features gives insights into the distribution of a single feature (cf. Section \ref{sect:varfract}), classification separates classes from each other, potentially in a non-linear fashion, which is a more sensitive way to detect differences between the text categories than a linear analysis of single properties. 
In this section, we describe the results for the classification of the text categories in detail. As indicated above (see Section \ref{sec:corpus}), we distinguish two classification tasks: Literary texts are classified against non-literary texts (Task 1), and canonical literary texts against non-canonical literary texts (Task 2). 
Moreover, for a better understanding of the postulated level of text processing, we present results for the low-level and high-level properties separately, as well as in combination (Table \ref{table:classification}). For classification, we used a Support Vector Machine (SVM) with a Radial Basis Function (RBF) kernel. As the features have varying scales, we normalized them to have a mean of 0 and standard deviation of 1. The evaluation measure is balanced accuracy, which is a weighted average accuracy value that is proportional to the size of each class and, therefore, does not favor larger classes. We assessed statistical significance of differences between settings by using a 5$\times$2cv paired $t$ test (\citealt{Dietterich1998}) (significance level at $p \leq 0.05$). In this test, 2-fold cross validation is repeated 5 times and the dataset is shuffled each time. In Table \ref{table:classification}, we report the mean and the standard deviation for the 10 runs for each setting.

\begin{table}
    \centering
    \caption{Accuracy of classification (in  \%) for the non-literary/literary distinction (Task 1) and the canonical/non-canonical distinction (Task 2). $Means \pm SD$ are listed ($N$ = 10). All values are significantly different ($p \leq 0.05$) from random accuracy (50\%), except where indicated by a dagger ($\dagger$).} 
    \begin{tabular}{c|ll|ll}
         & \multicolumn{2}{c|}{Task 1} & \multicolumn{2}{c}{Task 2} \Tstrut\\
         & \multicolumn{1}{c}{Variability} & \multicolumn{1}{c|}{Fractal Features} & \multicolumn{1}{c}{Variability} & \multicolumn{1}{c}{Fractal Features}\\\hline
        Noun & 71.0 $\pm$ 2.5 & 75.3 $\pm$ 2.3 & 69.5 $\pm$ 3.8 & 62.4 $\pm$ 3.0 \Tstrut\\
        Verb & 56.8 $\pm$ 3.7 & 75.1 $\pm$ 1.5 & 68.3 $\pm$ 2.1 & 55.5 $\pm$ 3.0 \\
        Adjective & 74.1 $\pm$ 2.7 & 80.4 $\pm$ 2.3 & 69.7 $\pm$ 4.0 & 51.6 $\pm$ 3.7$^{\dagger}$ \\
        Pronoun & 69.5 $\pm$ 0.9 & 72.1 $\pm$ 1.8 & 68.0 $\pm$ 1.9 & 52.2 $\pm$ 4.7$^{\dagger}$ \\
        Sentence-Length & 65.0 $\pm$ 2.2 & 74.0 $\pm$ 2.0 & 69.3 $\pm$ 2.9 & 59.7 $\pm$ 3.2 \\
        MTLD & 63.7 $\pm$ 2.3 & 56.9 $\pm$ 3.2 & 52.3 $\pm$ 3.3$^{\dagger}$ & 55.5 $\pm$ 3.1 \Tstrut\\
        Topic Distribution & 62.8 $\pm$ 2.3 & 64.0 $\pm$ 3.3 & 60.6 $\pm$ 3.4 & 49.2 $\pm$ 3.5$^{\dagger}$ \Bstrut\\\hline\hline
        Low-Level & 92.4 $\pm$ 2.1 & 86.0 $\pm$ 2.0 & 71.6 $\pm$ 2.6 & 62.9 $\pm$ 3.9 \Tstrut\\ 
        Low-Level, Combined & \multicolumn{2}{c|}{94.9 $\pm$ 1.0} & \multicolumn{2}{c}{71.4 $\pm$ 4.8} \Bstrut\\
        High-Level & 72.4 $\pm$ 1.9 & 63.2 $\pm$ 3.3 & 63.5 $\pm$ 3.2 & 57.1 $\pm$ 1.8 \Tstrut\\ 
        High-Level, Combined & \multicolumn{2}{c|}{71.8 $\pm$ 2.9} & \multicolumn{2}{c}{61.9 $\pm$ 4.5} \Bstrut\\\hline\hline
        Low- \& High-Level & 93.6 $\pm$ 1.3 & 84.9 $\pm$ 1.6 & 73.6 $\pm$ 2.3 & 65.0 $\pm$ 1.7 \Tstrut\\
        Low- \& High-Level, Combined & \multicolumn{2}{c|}{94.7 $\pm$ 1.3} & \multicolumn{2}{c}{71.6 $\pm$ 3.6} \Bstrut\\\hline
    \end{tabular}
    \label{table:classification}
\end{table}

The top part of Table \ref{table:classification} shows the classification results for individual properties. On the one hand, the analysis of variability provides comparable accuracies in Task 1 and Task 2. Exceptions are provided by verb frequency, which leads to much higher classification rates in Task 2 than in Task 1, and MTLD values, which are a better predictor in Task 1. The best performance is observed for adjective frequency, which yields the highest accuracy of all predictors in Task 1 and provides the best results in Task 2 as well (see also Table \ref{table:confidence-interval}). The variance of MTLD values is more powerful in distinguishing literary texts from non-literary text (Task 1), but it cannot separate canonical from non-canonical texts in Task 2.  As a lexical diversity measure, MTLD reflects the richness of vocabulary of a text. To get a better understanding of lexical diversity of literary and non-literary texts, we submitted the global MTLD-values of the texts, grouped into the categories `non-literary', `literary/canonical' and `literary/non-canonical', to an ANOVA.
The test did not reveal a significant difference between the lexical diversity of the text categories (p=0.68).
This finding is surprising, as lexical diversity is often regarded as a hallmark of good authorship,
and can thus be expected to vary across the sub-corpora of interest.

The fractal features result in better accuracies in Task 1 than in Task 2 for all properties, with the exception of MTLD, which performs similarly in both tasks. The highest classification rate for Task 1 is, again, obtained for adjective time series (80.4\%). The time series of low-level properties, i.e. POS-tags frequencies and sentence length, perform well in Task 1. By contrast, the fractal features cannot distinguish well between canonical and non-canonical literary texts (Task 2). This result is in accordance with the finding that the degree of fractality ($\mathcal{H}$) and the degree of asymmetry ($\mathcal{A}$) are of similar magnitude for canonical and non-canonical texts for almost all text properties (cf. Table \ref{table:confidence-interval}).

The POS-tag frequencies and sentence length are regarded as low-level properties and MTLD and topic distribution as high-level properties. The top part of Table \ref{table:classification} presents the classification results for variance and the fractal features separately. When combining the two features for all low-level and all high-level properties, respectively, as shown in the middle part of the table, a considerably improved accuracy is achieved in Task 1. Although the variance of each property alone does not provide a classification accuracy higher than 74\% (for adjective), their combination effectively raises the accuracy up to 92\%. 
Using all fractal features together for the classification task also increases the performance considerably. Finally, when all variances and fractal features are combined, the performance gets even better. Applying 5$\times$2cv paired $t$ test confirms that all of these improvements are significant. In Task 2, we do not observe such a large improvement in accumulating the variances or the fractal features. For example, the performance of a model combining all variances of low-level features is only slightly better than the performance of the variance of noun or adjective frequencies. For the fractal features, the classification accuracy of the combined model is similar to that of noun time series only. The combination of all features does not offer any improvement either.  

Similarly, we ran the classification task using all high-level properties. In Task 1 (cf. the middle part of Table 2), the combination of the variances of two high-level properties results in a considerable improvement. In contrast, the combination of the fractal features offers no enhancement in the result and is statistically similar to the performance using topic distribution. It is therefore expected that the combination of all variances and fractal features does not improve classification. Adding more features to an SVM classifier may actually decrease the classification result, because the SVM classifier tries to maximize generalization. Such a decrease is observed when all features are combined together.
In Task 2, we observe that the combination of variances of the high-level features improves the classification results, though not for the fractal features. The accumulation of all features does not provide any obvious improvement either. 

Low-level and high-level properties can be combined to analyze the different classes of text, as shown at the bottom of Table \ref{table:classification}. In Task 1, we observe no improvement when combining all variances or all fractal features. Finally, the result obtained by combining all features is not significantly different from the classifier that was trained on all features (variances and fractal features) of low-level properties. In Task 2, when all variances or all fractal features are taken into account, an improvement can be observed. The combination of all features does not improve the accuracy of the model compared to the model trained on all variances.

In summary, the results of the classification experiment show that low-level properties are more effective in distinguishing literary text from non-literary text (Task 1) than high-level properties. Even individual properties -- the frequencies of nouns and verbs -- reach accuracies higher than $70\%$, or even $80\%$ in the case of the fractal features for adjectives. By combining low-level features in the classification task, the accuracy reaches 95\%. The accuracy values for Task 2 range between $68-70\%$ for individual low-level features, and are much lower for high-level features. The performance of the classifier does not improve significantly if the low-level features are combined, and the resulting accuracy score ($71.6\%$) is not significantly better than the score for adjective frequencies ($69.7\%$). This finding points to a strong correlation of the low-level features in Task 2.

%%%%%%%%%%%%%%%%%%%%%%%%%%%%%%%%%%%%%%%%%%%%%%%%%%%%%%%%%%%%%%%%%%%%%%%%%%
%%%%%%%%%%%%%%%%%%%%%%%%%%%%%%%%%%%%%%%%%%%%%%%%%%%%%%%%%%%%%%%%%%%%%%%%%%
\section{Discussion and Conclusions}\label{sec:discussion}

What makes literature aesthetically pleasing? While many researchers have tackled this question by studying the semantics and cultural context of literary text, we focus on the formal structure of text in this study, an approach taken less commonly (cf. Section \ref{sec:introduction}). Specifically, we put forward ideas to generate time series of formal structural text properties (Section \ref{sec:text-representation}), and to study the global properties of these time series across individual texts (Section \ref{sec:methods}). Previous research in visual aesthetics suggests that global features of stimuli, such as fractality and variability, are associated with the aesthetic preference of human observers (cf. Section \ref{sec:introduction}). Here, we propose to analyze similar global features in texts of varying aesthetic claim or prestige (non-literary, literary/canonical and literary/non-canonical texts). 
For selected text properties that seemed particularly promising to us, we carried out a pilot study to validate 
our approach towards the analysis of objective (measurable) global features that vary across the three text categories. Such features will be considered as candidate predictors of textual aesthetics. 
The results of our analyses are briefly summarized in the following subsections.

\subsection{Classification of Text Types}

We analyzed three subcategories of text, associated with different degrees of aesthetic claim (canonical $>$ non-canonical $>$ non-literary). An ANOVA revealed differences between the three subcategories for several of the properties and features (Table \ref{table:confidence-interval}). In addition, we considered two specific tasks for a classification experiment. In Task 1, we compared the two literary text categories (canonical and non-canonical) with the non-literary category and in Task 2, the canonical texts with the non-canonical texts (see Sections \ref{sec:introduction} and \ref{sec:corpus}). We assumed that global structural features of literary texts are more similar between canonical and non-canonical texts, which would make Task 2 more difficult. 

As we expected, Task 1 resulted in higher classification accuracies than Task 2 in general (Table 2). Also, our results show that variability is more effective in classifying canonical, non-canonical, and non-literary texts than the fractal features. Moreover, fractal analysis is less successful in Task 2 than in Task 1. For the variances, classification rates reached a maximum rate of $94\%$ in Task 1 for a combination of all variances, and of $74\%$ in Task 2, respectively. This finding conforms to our expectation that Task 2 is more difficult than Task 1. For the fractal features, the Hurst exponent, $\mathcal{H}$, for the frequencies of verbs, adjectives and pronouns (as well as for MTLD) is lower for literary text than for non-literary text (Task 1), and literary texts display more fractal symmetry than non-literary text (Task 1), for all low-level properties. In Task 2, the fractal features do not result in a good separation of canonical vs. non-canonical texts. Accordingly, the $\mathcal{H}$ values and the asymmetry values ($\mathcal{A}$) do not differ much between the two literary categories for any of the text properties analyzed (Table 1). We thus conclude that canonical and non-canonical texts differ less in their global structural features than literary and non-literary texts.

\subsection{Low-Level and High-Level Properties}
The present findings show that, in general, the structural analysis of low-level properties (POS-tag frequencies and sentence length) results in a better separation of the three text categories than the analysis of high-level properties (MTLD and topic distribution). 
Regarding MTLD, it is important to note that lexical diversity is often regarded as a hallmark of good authorship. However, our analyses showed that literary texts do not make use of a broader range of vocabulary than non-literary texts; neither has such difference been observed for canonical as opposed to non-canonical texts.
Considering the high-level properties, only one of the four features studied, the variance $\mathcal{V}$, showed differences between the text categories. No differences were observed for any of the fractal features, except for the Hurst exponents for literary and non-literary texts (cf. Table \ref{table:confidence-interval}). Accordingly, classification rates obtained by using the high-level properties are relatively low (up to $64\%$; cf. Table \ref{table:classification}). By contrast, for the low-level properties, we observe differences between the text categories, both in terms of their variances and their fractal features. Low-level properties yield higher classification rates both individually (up to $80\%$; cf. Table \ref{table:classification}) and in combination (up to $95\%$).

At first glance it may seem surprising that the global distribution of low-level properties leads to a better separation of the text categories than the high-level properties. However, low-level properties have been associated with aesthetic preference in other sensory domains as well. In the visual domain, the global spatial distribution of several low-level properties (for example, luminance changes, edge orientations, curvilinear shape and color features; see Section \ref{sec:introduction}) has been related to the global structure of traditional artworks and other preferred visual stimuli. In the auditory domain, music has been shown to be characterized by fluctuations in low-level features, such as loudness and pitch \citep{voss1975}, frequency intervals \citep{Hsu1991}, sound amplitude \citep{Kello2017, roeske2018},
and other simple metrices, such as measures of pitch, duration, melodic intervals and harmonic intervals \citep{Manaris2005},
as well as patterns of consonance \citep{Wu2015}.
These and many other studies indicate that low-level properties of music show long-range correlations that are scale-invariant and obey a power law.
Interestingly, similar results were obtained for animal song \citep{Kello2017, roeske2018}.

We surmise that low-level properties of text primarily reflect discourse modes \citep{smith2003}. These modes -- Narrative, Report, Description (temporal), Information and Argument (atemporal)  -- are associated with different frequency distributions of POS-tags (cf. also \citealt{Biber1995}, who uses more specific categories in his multi-dimensional register analysis, however). For example, the  Narrative mode is associated with verbs, while Description requires more adjectives. In a comparison of literary and non-literary text, it is moreover important to bear in mind that literary text implies both external communication (between the narrator and the reader) and internal communication (between the protagonists, in the form of dialogues) as well as internal monologues and thoughts. Our results suggest that non-literary texts show more global variability between discourse modes than literary texts, while the time series are smoother, pointing to more local homogeneity (clearer structural differentiation as reflected in hierarchical text structure). Canonical literary texts seem to pattern with non-literary texts in terms of their higher global variability, in comparison to non-canonical
literature. While this hypothesis requires more (qualitative) in-depth studies, it suggests that canonical authors may use
a richer variety of discourse modes than non-canonical authors.

\subsection{Variability Versus Fractal Features}

Our results show that global variability, operationalized as variance, is an important feature that distinguishes canonical from non-canonical texts. In general, the variability of canonical texts is higher than the variability of non-canonical texts, for all properties investigated by us. The pattern concerning the variability of non-literary texts in comparison to literary texts is less uniform. For most properties, variability is higher for non-literary than for literary texts. As a result, the variability of canonical texts is closer to (or the same as) that of non-literary texts. Only for pronoun frequencies and MTLD values can a different pattern be observed. Here, canonical texts are more variable than both non-canonical and non-literary texts.

A direct comparison with the visual domain is difficult due to the different dimensions and study paradigms used in visual and literary studies. Nevertheless, the formal structure of traditional visual artworks was also described as particularly rich and variable, compared to other types of natural and man-made objects and scenes \citep{brachmann-redies2017, redies-brachmann2017}.

MFDFA, the method used by us to detect multi-fractal patterns, has shown that long-range correlations are less pronounced in literary texts than in non-literary texts, for all properties except sentence length and topic distributions. Moreover, the similarity of the Hurst exponent, $\mathcal{H}$, of all text properties for canonical and non-canonical texts suggests that a particular degree of fractality could be a universal characteristic of different categories of literary texts, regardless of the specific text category. The fact that non-literary texts seem to exhibit a higher degree of fractality than literary ones could be related to the smoother signal, reflecting local homogeneity (in terms of discourse modes), as pointed out above.

In the visual domain, traditional artworks can be characterized by an intermediate to high degree of self-similarity \citep{Braun2013, brachmann-redies2017}.
In the Fourier domain, large subsets of traditional artworks have spectral properties similar to pink noise, with a power ($1/f^p$) spectral exponent around $p = 1$ \citep{graham2007, redies2007}, which is also characteristic of many (but not all) natural patterns and scenes \citep{Tolhurst1992}.
In MFDFA, this corresponds to a Hurst exponent of $1$, while $\mathcal{H} = 0.5$ indicates white noise (no long-range correlations; corresponding to a Fourier power spectral exponent of $0$). The median $\mathcal{H}$ value for the different text properties ranges from $0.63$ to $0.73$ in our study, confirming previous results for sentence length by \citep{Drozdz2016}.
This degree of self-similarity thus lies in between that of most natural signals and random (white) noise. The relevance of this finding requires further exploration. In particular, detailed qualitative studies of individual texts will be required to gain a better understanding of the distributional factors giving rise to fractal patterns.

\subsection{Generalization and Future Directions}

While our study has led to some non-trivial results, it has some obvious limitations. First, we investigated English texts only, and these texts were taken from a restricted time period (19th and early 20th centuries). In order to investigate whether any of the present findings can be generalized to other types of literary texts, other languages and other time periods would have to be investigated separately. If variability and fractality do not represent universal characteristics of literary texts, they might still be useful to distinguish different literary styles or genres, and perhaps texts from different epochs or cultures.

Our sample of texts has been deliberately limited to prose. We did not study poetry, for several reasons. First, several of the features analyzed in the present study require long texts, which are uncommon in poetry. Studying fractality of time series is possible if they are long enough to potentially exhibit long-range correlation. To study poetry, our analysis tools would thus have to be adapted to short texts. Second, the aesthetics of poetry can only be studied in relation to metre and other prodosic properties (e.g. rhyme), and thus requires an entirely different approach.

A number of questions for future studies arise from a medical point of view. Does brain damage have an impact on patterns of variability and fractality in texts? Such effects have been observed in the visual domain. For example, dementia and cerebral stroke have been shown to alter artistic creativity \citep{Miller1998, Sherwood2012}.
Changes in low-level image properties have also been observed
in the art produced by persons with schizophrenia,  \citep{henemann2017}.

Our results suggest that high-level properties are less distinctive than low-level properties for distinguishing both literary as opposed to non-literary texts, and canonical as opposed to non-canonical texts. However, it is obviously possible that there are other ways of operationalizing and measuring high-level features that we have not taken into consideration. Identifying other methods that reflect propositional information and comprehension is thus an important task for future studies. In the visual domain, it is generally agreed that both the formal properties as well as the content and context of aesthetical stimuli can contribute to their liking by human observers \citep{Chatterjee2014, redies2015}.

Another limitation is that, while we have provided a broad list of possible text properties and global features, we did not study all of their combinations, due to limitations of space and time. Instead, we focused on a set of properties and features that seemed particularly promising to us. Indeed, the low-level properties that we selected proved efficient in the classification tasks. However, using high-level properties, such as MTLD and topic distribution did not yield high classification rates. Studying other text properties and their combination is thus another important future line of research.

%%%%%%%%%%%%%%%%%%%%%%%%%%%%%%%%%%%%%%%%%%%%%%%%%%%%%%%%%%%%%%%%%%%%%%%%%%
%%%%%%%%%%%%%%%%%%%%%%%%%%%%%%%%%%%%%%%%%%%%%%%%%%%%%%%%%%%%%%%%%%%%%%%%%%
%%%%%%%%%%%%%%%%%%%%%%%%%%%%%%%%%%%%%%%%%%%%%%%%%%%%%%%%%%%%%%%%%%%%%%%%%%

\bibliographystyle{frontiersinSCNS_ENG_HUMS} % for Science, Engineering and Humanities and Social Sciences articles, for Humanities and Social Sciences articles please include page numbers in the in-text citations
\bibliography{refs}

%%% Make sure to upload the bib file along with the tex file and PDF
%%% Please see the test.bib file for some examples of references

%%% If you are submitting a figure with subfigures please combine these into one image file with part labels integrated.
%%% If you don't add the figures in the LaTeX files, please upload them when submitting the article.
%%% Frontiers will add the figures at the end of the provisional pdf automatically
%%% The use of LaTeX coding to draw Diagrams/Figures/Structures should be avoided. They should be external callouts including graphics.

\end{document}

\onecolumn
\firstpage{1}

\title[Supplementary Material]{{\helveticaitalic{Supplementary Material}}}

\maketitle

\fontsize{7}{9}\selectfont

\begin{longtable}{|l|p{8cm}|p{4cm}|l|}
\caption{List of texts in the Jena Textual Aesthetics Corpus. Canonical texts were selected from the Corpus of Canonical Western Literature. Non-canonical texts were downloaded from www.smashwords.com, www.goodreads.com, www.feedbooks.com, or Project Gutenberg. Non-literary texts were sampled from Project Gutenberg.}\\
\hline
\textbf{} & \textbf{Title} & \textbf{Author(s)} & \textbf{Category} \\
 \hline
\endfirsthead
\multicolumn{4}{c}%
{\tablename\ \thetable\ -- \textit{Continued from previous page}} \\
\hline
\textbf{} & \textbf{Title} & \textbf{Author(s)} & \textbf{Category} \\
\hline
\endhead
\hline \multicolumn{4}{r}{\textit{Continued on next page}} \\
\endfoot
\hline
\endlastfoot

1 & Little Dorrit & Charles Dickens & Canonical \\ \hline
2 & Oliver Twist & Charles Dickens & Canonical \\ \hline
3 & The Life and Adventures of Nicholas Nickleby & Charles Dickens & Canonical \\ \hline
4 & The Mystery of Edwin Drood & Charles Dickens & Canonical \\ \hline
5 & The Pickwick Papers & Charles Dickens & Canonical \\ \hline
6 & Jane Eyre & Charlotte Bronte & Canonical \\ \hline
7 & Villette & Charlotte Bronte & Canonical \\ \hline
8 & Cranford & Elizabeth Gaskell & Canonical \\ \hline
9 & Mary Barton & Elizabeth Gaskell & Canonical \\ \hline
10 & North and South & Elizabeth Gaskell & Canonical \\ \hline
11 & Agnes Grey & Anne Bronte & Canonical \\ \hline
12 & Adam Bede & George Eliot & Canonical \\ \hline
13 & Daniel Deronda & George Eliot & Canonical \\ \hline
14 & Middlemarch & George Eliot & Canonical \\ \hline
15 & Silas Marner & George Eliot & Canonical \\ \hline
16 & The Mill on the Floss & George Eliot & Canonical \\ \hline
17 & Emma & Jane Austen & Canonical \\ \hline
18 & Mansfield Park & Jane Austen & Canonical \\ \hline
19 & Persuasion & Jane Austen & Canonical \\ \hline
20 & Pride and Prejudice & Jane Austen & Canonical \\ \hline
21 & The Picture of Dorian Gray & Oscar Wilde & Canonical \\ \hline
22 & The Tenant of Wildfell Hall & Anne Bronte & Canonical \\ \hline
23 & Sartor Resartus & Thomas Carlyle & Canonical \\ \hline
24 & Old Mortality & Walter Scott & Canonical \\ \hline
25 & Redgauntlet & Walter Scott & Canonical \\ \hline
26 & The Heart of Midlothian & Walter Scott & Canonical \\ \hline
27 & Waverley & Walter Scott & Canonical \\ \hline
28 & No Name & Wilkie Collins & Canonical \\ \hline
29 & The Moonstone & Wilkie Collins & Canonical \\ \hline
30 & The Woman in White & Wilkie Collins & Canonical \\ \hline
31 & The History of Henry Esmond & William Makepeace Thackeray & Canonical \\ \hline
32 & Vanity Fair & William Makepeace Thackeray & Canonical \\ \hline
33 & Dracula & Bram Stoker & Canonical \\ \hline
34 & The Well at the World's end & William Morris & Canonical \\ \hline
35 & The Narrative of Arthur Gordon Pym & Edgar Allan Poe & Canonical \\ \hline
36 & The Ambassadors & Henry James & Canonical \\ \hline
37 & The Awkward Age & Henry James & Canonical \\ \hline
38 & The Bostonians & Henry James & Canonical \\ \hline
39 & The Golden Bowl & Henry James & Canonical \\ \hline
40 & The Portrait of a Lady & Henry James & Canonical \\ \hline
41 & The Wings of Dove & Henry James & Canonical \\ \hline
42 & Moby Dick & Herman Melville & Canonical \\ \hline
43 & The Deerslayers & James Fenimore Cooper & Canonical \\ \hline
44 & A Christmas Carol & Charles Dickens & Canonical \\ \hline
45 & Little Women & Louisa May Alcott & Canonical \\ \hline
46 & Puddnhead Wilson & Mark Twain & Canonical \\ \hline
47 & The Adventures of Finn & Mark Twain & Canonical \\ \hline
48 & The Mysterious Stranger & Mark Twain & Canonical \\ \hline
49 & The Marble Faun & Nathaniel Hawthorne & Canonical \\ \hline
50 & The Scarlet Letter & Nathaniel Hawthorne & Canonical \\ \hline
51 & The Education of Adams & Henry Adams & Canonical \\ \hline
52 & Walden & Henry David Thoreau & Canonical \\ \hline
53 & A Conneticut Yankee in King Arthurs & Mark Twain & Canonical \\ \hline
54 & Babbitt & Sinclair Lewis & Canonical \\ \hline
55 & A Tale of Two Cities & Charles Dickens & Canonical \\ \hline
56 & Sister Carrie & Theodore Dreiser & Canonical \\ \hline
57 & My Antonia & Willa Cather & Canonical \\ \hline
58 & The Old Wives Tale & Arnold Bennett & Canonical \\ \hline
59 & Portrait of the Artist as a Young Man & James Joyce & Canonical \\ \hline
60 & Ulysses & James Joyce & Canonical \\ \hline
61 & Lord Jim & Joseph Conrad & Canonical \\ \hline
62 &  Nostromo & Joseph Conrad & Canonical \\ \hline
63 & The Secret Agent & Joseph Conrad & Canonical \\ \hline
64 & Under Western Eyes & Joseph Conrad & Canonical \\ \hline
65 &  Victory: An Island Tale & Joseph Conrad & Canonical \\ \hline
66 & Bleak House & Charles Dickens & Canonical \\ \hline
67 & The Rainbow & Lawrence D.H & Canonical \\ \hline
68 & Women in Love & Lawrence D.H & Canonical \\ \hline
69 &  Kim & Rudyard Kipling & Canonical \\ \hline
70 & Puck of Pooks Hill & Rudyard Kipling & Canonical \\ \hline
71 & Jude the Obscure & Thomas Hardy & Canonical \\ \hline
72 & Tess of the dUrbervilles & Thomas Hardy & Canonical \\ \hline
73 & The Mayor of Casterbridge & Thomas Hardy & Canonical \\ \hline
74 & The Return of the Native & Thomas Hardy & Canonical \\ \hline
75 & David Copperfield & Charles Dickens & Canonical \\ \hline
76 & Great Expectations & Charles Dickens & Canonical \\ \hline
77 & Hard Times & Charles Dickens & Canonical \\ \hline
78 & The Face in the Abyss & Abraham Merritt & Non-Canonical \\ \hline
79 & A Prisoner in Fairyland & Algernon Blackwood & Non-Canonical \\ \hline
80 & The Centaur & Algernon Blackwood & Non-Canonical \\ \hline
81 & Ruth Fielding at the War Front & Alice B. Emerson & Non-Canonical \\ \hline
82 & The International Spy & Allen Upward & Non-Canonical \\ \hline
83 & A Texas Matchmaker & Andy Adams & Non-Canonical \\ \hline
84 & The Filigree Ball & Anna Katharine Green & Non-Canonical \\ \hline
85 & Looking Further Backward & Arthur Dudley Vinton & Non-Canonical \\ \hline
86 & The Hill Of Dreams & Arthur Machen & Non-Canonical \\ \hline
87 & The Elusive Pimpernel & Baroness Emma Orczy & Non-Canonical \\ \hline
88 & The Gloved Hand & Burton E. Stevenson & Non-Canonical \\ \hline
89 & Jean of the Lazy A & B.M . Bower & Non-Canonical \\ \hline
90 & Wieland : or , The Transformation & Charles Brockden Brown & Non-Canonical \\ \hline
91 & The Great Quest & Charles Hawes & Non-Canonical \\ \hline
92 & The Filibusters & Charles John Cutcliffe Wright Hyne & Non-Canonical \\ \hline
93 & Bar-20 Days & Clarence E. Mulford & Non-Canonical \\ \hline
94 & Wunpost & Dane Coolidge & Non-Canonical \\ \hline
95 & The Girl of the Golden West & David Belasco & Non-Canonical \\ \hline
96 & Love Insurance & Earl Derr Biggers & Non-Canonical \\ \hline
97 & The Wouldbegoods & Edith Nesbit & Non-Canonical \\ \hline
98 & Wet Magic & Edith Nesbit & Non-Canonical \\ \hline
99 & Philip Dru : Administrator & Edward Mandell House & Non-Canonical \\ \hline
100 & An Amiable Charlatan & Edward Phillips Oppenheim & Non-Canonical \\ \hline
101 & The Double Traitor & Edward Phillips Oppenheim & Non-Canonical \\ \hline
102 & The Zeppelin 's Passenger & Edward Phillips Oppenheim & Non-Canonical \\ \hline
103 & The People of the Ruins & Edward Shanks & Non-Canonical \\ \hline
104 & The Honor of the Name & mile Gaboriau & Non-Canonical \\ \hline
105 & Kai Lung's Golden Hours & Ernest Bramah Smith & Non-Canonical \\ \hline
106 & The Riddle of the Sands & Erskine Childers & Non-Canonical \\ \hline
107 & The Missourian & Eugene Percy Lyle & Non-Canonical \\ \hline
108 & Privy Seal & Ford Madox Ford & Non-Canonical \\ \hline
109 & The Ivory Snuff Box & Frederic Arnold Kummer & Non-Canonical \\ \hline
110 & The Afterglow & George Allan England & Non-Canonical \\ \hline
111 & The Flying Legion & George Allan England & Non-Canonical \\ \hline
112 & West Wind Drift & George Barr McCutcheon & Non-Canonical \\ \hline
113 & Peter the Brazen & George F. Worts & Non-Canonical \\ \hline
114 & Olga Romanoff or , The Syren of the Skies & George Griffith & Non-Canonical \\ \hline
115 & The Princess and Curdie & George MacDonald & Non-Canonical \\ \hline
116 & The Adventures of Don Lavington & George Manville Fenn & Non-Canonical \\ \hline
117 & A Voyage to the Moon & George Tucker & Non-Canonical \\ \hline
118 & Claim Number One & George W. Ogden & Non-Canonical \\ \hline
119 & The Flockmaster of Poison Creek & George W. Ogden & Non-Canonical \\ \hline
120 & Trilby & George du Maurier & Non-Canonical \\ \hline
121 & Rose O'Paradise & Grace Miller White & Non-Canonical \\ \hline
122 & Condemned as a Nihilist & G. A. Henty & Non-Canonical \\ \hline
123 & Man on the Box & Harold MacGrath & Non-Canonical \\ \hline
124 & The Puppet Crown & Harold MacGrath & Non-Canonical \\ \hline
125 & The Blind Spot & Homer Eon Flint & Non-Canonical \\ \hline
126 & Men of Iron & Howard Pyle & Non-Canonical \\ \hline
127 & The Dark House & Ida Alexa Ross Wylie & Non-Canonical \\ \hline
128 & The Daughter of Brahma & Ida Alexa Ross Wylie & Non-Canonical \\ \hline
129 & Towards Morning & Ida Alexa Ross Wylie & Non-Canonical \\ \hline
130 & Jurgen : A Comedy of Justice & James Branch Cabell & Non-Canonical \\ \hline
131 & A Strange Manuscript Found in a Copper Cylinder & James De Mille & Non-Canonical \\ \hline
132 & Lost in the Fog & James De Mille & Non-Canonical \\ \hline
133 & Varney the Vampire & James Malcom Rymer & Non-Canonical \\ \hline
134 & The Danger Trail & James Oliver Curwood & Non-Canonical \\ \hline
135 & The Lost Stradivarius & John Meade Falkner & Non-Canonical \\ \hline
136 & The Nebuly Coat & John Meade Falkner & Non-Canonical \\ \hline
137 & The Weapons of Mystery & Joseph Hocking & Non-Canonical \\ \hline
138 & The Chestermarke Instinct & Joseph Smith Fletcher & Non-Canonical \\ \hline
139 & Afloat On The Flood & Lawrence J. Leslie & Non-Canonical \\ \hline
140 & Diane of the Green Van & Leona Dalrymple & Non-Canonical \\ \hline
141 & Don Rodriguez : Chronicles of Shadow Valley & Lord Dunsany & Non-Canonical \\ \hline
142 & The Treasure Trail & Marah Ellis Ryan & Non-Canonical \\ \hline
143 & Mizora : A Prophecy & Mary E. Bradley & Non-Canonical \\ \hline
144 & Dangerous Days & Mary Roberts Rinehart & Non-Canonical \\ \hline
145 & The Blue Germ & Maurice Nicoll & Non-Canonical \\ \hline
146 & The Night Horseman & Max Brand & Non-Canonical \\ \hline
147 & The Sleuth of St. James 's Square & Melville Davisson Post & Non-Canonical \\ \hline
148 & Across the Zodiac & Percy Greg & Non-Canonical \\ \hline
149 & Bardelys the Magnificent & Rafael Sabatini & Non-Canonical \\ \hline
150 & Soldiers of Fortune & Richard Harding Davis & Non-Canonical \\ \hline
151 & The Beetle & Richard Marsh & Non-Canonical \\ \hline
152 & The Triumphs of Eugne Valmont & Robert Barr & Non-Canonical \\ \hline
153 & Dawn of All & Robert Hugh Benson & Non-Canonical \\ \hline
154 & Erling the Bold & Robert Michael Ballantyne & Non-Canonical \\ \hline
155 & The Dog Crusoe and His Master & Robert Michael Ballantyne & Non-Canonical \\ \hline
156 & Ailsa Paige & Robert William Chambers & Non-Canonical \\ \hline
157 & In Search of the Unknown & Robert William Chambers & Non-Canonical \\ \hline
158 & In the Quarter & Robert William Chambers & Non-Canonical \\ \hline
159 & Under the Ocean to the South Pole & Roy Rockwood & Non-Canonical \\ \hline
160 & Erewhon , or Over The Range & Samuel Butler & Non-Canonical \\ \hline
161 & The road to Frontenac & Samuel Merwin & Non-Canonical \\ \hline
162 & Brood of the Witch-Queen & Sax Rohmer & Non-Canonical \\ \hline
163 & The Revolt of Man & Sir Walter Besant & Non-Canonical \\ \hline
164 & The Brass Bottle & Thomas Anstey Guthrie & Non-Canonical \\ \hline
165 & The Stray Lamb & Thorne Smith & Non-Canonical \\ \hline
166 & The Doomsman & Van Tassel Sutphen & Non-Canonical \\ \hline
167 & The Song of the Lark & Willa Cather & Non-Canonical \\ \hline
168 & The Old Tobacco Shop & William Bowen & Non-Canonical \\ \hline
169 & The Boats of the 'Glen-Carrig ' & William Hope Hodgson & Non-Canonical \\ \hline
170 & Hushed Up! & William Le Queux & Non-Canonical \\ \hline
171 & The Border Legion & Zane Grey & Non-Canonical \\ \hline
172 & The Desert of Wheat & Zane Grey & Non-Canonical \\ \hline
173 & Scottish Cathedrals and Abbeys & Dugald Butler & Non-Literary \\ \hline
174 & A Text-Book of the History of Architecture: Seventh Edition, revised & A. D. F. Hamlin & Non-Literary \\ \hline
175 & Some Account of Gothic Architecture in Spain & George Edmund Street & Non-Literary \\ \hline
176 & Japanese Homes and Their Surroundings & Edward Sylvester Morse & Non-Literary \\ \hline
177 & The Architecture of Provence and the Riviera & David MacGibbon & Non-Literary \\ \hline
178 & Historic Ornament, Vol. 2: Treatise on decorative art and architectural ornament & James Ward & Non-Literary \\ \hline
179 & Military Architecture in England During the Middle Ages & A. Hamilton Thompson & Non-Literary \\ \hline
180 & How to Study Architecture & Charles H. Caffin & Non-Literary \\ \hline
181 & Cakes \& Ale: A Dissertation on Banquets Interspersed with Various Recipes, More or Less Original, and anecdotes, mainly veracious & Edward Spencer & Non-Literary \\ \hline
182 & Food and Flavor: A Gastronomic Guide to Health and Good Living & Henry T. Finck & Non-Literary \\ \hline
183 & A Concise Dictionary of Middle English from A.D. 1150 to 1580 & A. L. Mayhew, Walter William Skeat & Non-Literary \\ \hline
184 & A Dictionary of Slang, Cant, and Vulgar Words: Used at the Present Day in the Streets of London & John Camden Hotten & Non-Literary \\ \hline
185 & The Devil's Dictionary & Ambrose Bierce & Non-Literary \\ \hline
186 & The Encyclopedia Britannica Vol. 1 & University of Cambridge & Non-Literary \\ \hline
187 & The Encyclopedia Britannica Vol. 2 & University of Cambridge & Non-Literary \\ \hline
188 & Glossary of Chess terms & Gregory Zorzos & Non-Literary \\ \hline
189 & Through the Brazilian Wilderness & Roosevelt & Non-Literary \\ \hline
190 & Gold, Sport, and Coffee Planting in Mysore & Robert H. Elliot & Non-Literary \\ \hline
191 & The Economic Aspect of Geology & C. K. Leith & Non-Literary \\ \hline
192 & The Shores of the Adriatic: The Austrian Side, The Kustenlande, Istria, and Dalmatia & F. Hamilton Jackson & Non-Literary \\ \hline
193 & Island Life; Or, The Phenomena and Causes of Insular Faunas and Floras & Alfred Russel Wallace & Non-Literary \\ \hline
194 & Sea and Sardinia & D. H. Lawrence & Non-Literary \\ \hline
195 & Sketches from the Subject and Neighbour Lands of Venice & Edward A. Freeman & Non-Literary \\ \hline
196 & The Elements of Geology & William Harmon Norton & Non-Literary \\ \hline
197 & The Principles of Stratigraphical Geology & J. E. Marr & Non-Literary \\ \hline
198 & Fragments of Earth Lore: Sketches \& Addresses Geological and Geographical & James Geikie & Non-Literary \\ \hline
199 & Earth Features and Their Meaning: An Introduction to Geology for the Student and the General Reader & William Herbert Hobbs & Non-Literary \\ \hline
200 & The Common Law & Oliver Wendell Holmes & Non-Literary \\ \hline
201 & Babylonian and Assyrian Laws, Contracts and Letters & C. H. W. Johns & Non-Literary \\ \hline
202 & Putnam's Handy Law Book for the Layman & Albert Sidney Bolles & Non-Literary \\ \hline
203 & Marriage and Divorce Laws of the World & Hyacinthe Ringrose & Non-Literary \\ \hline
204 & The Law and the Poor & Edward Abbott, Sir Parry & Non-Literary \\ \hline
205 & International Law. A Treatise. Vol. 1: Peace. Second Edition & L. Oppenheim & Non-Literary \\ \hline
206 & International Law. A Treatise. Vol. 2: War and Neutrality. Second Edition & L. Oppenheim & Non-Literary \\ \hline
207 & International Law & George Grafton Wilson, George Fox Tucker & Non-Literary \\ \hline
208 & The Criminal Prosecution and Capital Punishment of Animals & E. P. Evans & Non-Literary \\ \hline
209 & The English Constitution & Walter Bagehot & Non-Literary \\ \hline
210 & The Law of the Sea: A Manual of the Principles of Admiralty Law for Students, Mariners, and Ship Operators & George L. Canfield, George W. Dalzell, J. Y. Brinton & Non-Literary \\ \hline
211 & Woman and the Republic: A Survey of the Woman-Suffrage Movement in the United States and a Discussion of the Claims and Arguments of Its Foremost Advocates & Helen Kendrick Johnson & Non-Literary \\ \hline
212 & The American Judiciary & Simeon E. Baldwin & Non-Literary \\ \hline
213 & The Story of Evolution & Joseph McCabe & Non-Literary \\ \hline
214 & A Practical Physiology: A Text-Book for Higher Schools & Albert F. Blaisdell & Non-Literary \\ \hline
215 & Our Vanishing Wild Life: Its Extermination and Preservation & William T. Hornaday & Non-Literary \\ \hline
216 & Amusements in Mathematics & Henry Ernest Dudeney & Non-Literary \\ \hline
217 & On the Genesis of Species & St. George Jackson Mivart & Non-Literary \\ \hline
218 & An Elementary Study of Chemistry & William McPherson, William Edwards Henderson & Non-Literary \\ \hline
219 & Great Astronomers & Robert S. Ball & Non-Literary \\ \hline
220 & Evolution, Old \& New & Samuel Butler & Non-Literary \\ \hline
221 & Darwin, and After Darwin, Vol. 1: An Exposition of the Darwinian Theory and a Discussion of Post-Darwinian Questions & George John Romanes & Non-Literary \\ \hline
222 & Creative Evolution & Henri Bergson & Non-Literary \\ \hline
223 & Myths and Marvels of Astronomy & Richard A. Proctor & Non-Literary \\ \hline
224 & A Popular History of Astronomy During the Nineteenth Century: Fourth Edition & Agnes M. Clerke & Non-Literary \\ \hline
225 & Pioneers of Science & Oliver, Sir Lodge & Non-Literary \\ \hline
226 & A Text-Book of Astronomy & George C. Comstock & Non-Literary \\ \hline
227 & Astronomical Myths: Based on Flammarions's ``History of the Heavens" &  Camille Flammarion, J. F. Blake & Non-Literary \\ \hline
228 & Darwin, and After Darwin, Vol. 2: Post-Darwinian Questions, Heredity and Utility & George John Romanes & Non-Literary \\ \hline
229 & Astronomy: The Science of the Heavenly Bodies & David P. Todd & Non-Literary \\ \hline
230 & The Foundations of Science: Science and Hypothesis, The Value of Science, Science and Method & Henri Poincar\'e & Non-Literary \\ \hline
231 & A Civic Biology, Presented in Problems & George W. Hunter & Non-Literary \\ \hline
232 & Physics & Willis E. Tower, Charles M. Turton, Charles H. Smith, Thomas D. Cope & Non-Literary \\ \hline
233 & A Century of Science, and Other Essays & John Fiske & Non-Literary \\ \hline
234 & Side-Lights on Astronomy and Kindred Fields of Popular Science & Simon Newcomb & Non-Literary \\ \hline
235 & Elementary Zoology, Second Edition & Vernon L. Kellogg & Non-Literary \\ \hline
236 & Experiments on Animals & Stephen Paget & Non-Literary \\ \hline
237 & The Sea-beach at Ebb-tide: A Guide to the Study of the Seaweeds and the Lower Animal Life Found Between Tide-marks & Augusta Foote Arnold & Non-Literary \\ \hline
238 & The Making of Species & Douglas Dewar, Frank Finn & Non-Literary \\ \hline
239 & The Science and Philosophy of the Organism & Hans Driesch & Non-Literary \\ \hline
240 & Problems of Genetics & William Bateson & Non-Literary \\ \hline
241 & The Organism as a Whole, from a Physicochemical Viewpoint & Jacques Loeb & Non-Literary \\ \hline
242 & A Guide to the Study of Fishes, Vol. 1 & David Starr Jordan & Non-Literary \\ \hline
243 & Evolution: Its nature, its evidence, and its relation to religious thought & Joseph LeConte & Non-Literary \\ \hline
244 & The Races of Man: An Outline of Anthropology and Ethnography & Joseph Deniker & Non-Literary \\ \hline
245 & Physiology: The Science of the Body & Ernest G. Martin & Non-Literary \\ \hline
246 & Observations of a Naturalist in the Pacific Between 1896 and 1899, Vol. 1 & H. B. Guppy & Non-Literary \\ \hline
247 & Animal Life and Intelligence & C. Lloyd Morgan & Non-Literary \\ \hline
248 & A Guide to the Study of Fishes, Vol. 2 & David Starr Jordan & Non-Literary \\ \hline
249 & Stargazing: Past and Present & Norman, Sir Lockyer & Non-Literary \\ \hline
250 & Observations of a Naturalist in the Pacific Between 1896 and 1899, Vol. 2 & H. B. Guppy & Non-Literary \\ \hline
251 & Regeneration & Thomas Hunt Morgan & Non-Literary \\ \hline
252 & Telescopic Work for Starlight Evenings & William F. Denning & Non-Literary \\ \hline
253 & The Logic of Chance, 3rd edition & John Venn & Non-Literary \\ \hline
254 & Biology and Its Makers: With Portraits and Other Illustrations & William A. Locy & Non-Literary \\ \hline
255 & The Crayfish: An Introduction to the Study of Zoology & Thomas Henry Huxley & Non-Literary \\ \hline
256 & History of Botany (1530-1860) & Julius Sachs & Non-Literary \\ \hline
257 & The Universal Kinship & J. Howard Moore & Non-Literary \\ \hline
258 & The philosophy of biology & James Johnstone & Non-Literary \\ \hline
259 & Hygienic Physiology: with Special Reference to the Use of Alcoholic Drinks and Narcotics & Joel Dorman Steele & Non-Literary \\ \hline
260 & Species and Varieties, Their Origin by Mutation & Hugo de Vries & Non-Literary \\ \hline
261 & The Naturalist in La Plata & W. H. Hudson & Non-Literary \\ \hline
262 & Studies in the Psychology of Sex, Vol. 1 & Havelock Ellis & Non-Literary \\ \hline
263 & Studies in the Psychology of Sex, Vol. 2 & Havelock Ellis & Non-Literary \\ \hline
264 & The Mind of the Child, Part II: The Development of the Intellect & William T. Preyer & Non-Literary \\ \hline
265 & The Measurement of Intelligence & Lewis M. Terman & Non-Literary \\ \hline
266 & Human Traits and their Social Significance & Irwin Edman & Non-Literary \\ \hline
267 & Chapters in the History of the Insane in the British Isles & Daniel Hack Tuke & Non-Literary \\ \hline
268 & Human Personality and Its Survival of Bodily Death & F. W. H. Myers & Non-Literary \\ \hline
269 & Mysterious Psychic Forces: An Account of the Author's Investigations in Psychical Research, Together with Those of Other European Savants & Camille Flammarion & Non-Literary \\ \hline
270 & The Group Mind: A Sketch of the Principles of Collective Psychology & William McDougall & Non-Literary \\ \hline
271 & On the State of Lunacy and the Legal Provision for the Insane: With Observations on the Construction and Organization of Asylums & J. T. Arlidge & Non-Literary \\ \hline
272 & The Criminal & Havelock Ellis & Non-Literary \\ \hline
273 & Fact and Fable in Psychology & Joseph Jastrow & Non-Literary \\ \hline
274 & Mental Evolution in Man: Origin of Human Faculty & George John Romanes & Non-Literary \\ \hline
275 & A Beginner's Psychology & Edward Bradford Titchener & Non-Literary \\ \hline
276 & Mental diseases: A Public Health Problem & James Vance May & Non-Literary \\ \hline
277 & The Law of Psychic Phenomena & Thomson Jay Hudson & Non-Literary \\ \hline
278 & Psychology: Briefer Course & William James & Non-Literary \\ \hline
279 & The Principles of Psychology, Vol. 1 & William James & Non-Literary \\ \hline
280 & The Principles of Psychology, Vol. 2 & William James & Non-Literary \\ \hline
281 & Sex \& Character & Otto Weininger & Non-Literary \\ \hline
282 & Youth: Its Education, Regimen, and Hygiene & G. Stanley Hall & Non-Literary \\ \hline
283 & Ten Thousand Dreams Interpreted; Or, What's in a Dream: A Scientific and Practical Exposition & Gustavus Hindman Miller & Non-Literary \\ \hline
284 & Browning as a Philosophical and Religious Teacher & Henry, Sir Jones & Non-Literary \\ \hline
285 & The Life of Reason: The Phases of Human Progress & George Santayana & Non-Literary \\ \hline
286 & An Introduction to Philosophy & George Stuart Fullerton & Non-Literary \\ \hline
287 & The Approach to Philosophy & Ralph Barton Perry & Non-Literary \\ \hline
288 & The Will to Believe, and Other Essays in Popular Philosophy & William James & Non-Literary \\ \hline
289 & Christianity and Greek Philosophy & B. F. Cocker & Non-Literary \\ \hline
290 & A History of Mediaeval Jewish Philosophy & Isaac Husik & Non-Literary \\ \hline
291 & The Mediaeval Mind (Vol. 1 of 2): A History of the Development of Thought and Emotion in the Middle Ages & Henry Osborn Taylor & Non-Literary \\ \hline
292 & The Mediaeval Mind (Vol. 2 of 2): A History of the Development of Thought and Emotion in the Middle Ages & Henry Osborn Taylor & Non-Literary \\ \hline
293 & The Philosophy of Friedrich Nietzsche & H. L. Mencken & Non-Literary \\ \hline
294 & Philosophical Studies & G. E. Moore & Non-Literary \\ \hline
295 & What Nietzsche Taught & Willard Huntington Wright & Non-Literary \\ \hline
296 & The Greek Philosophers, Vol. 1 & Alfred William Benn & Non-Literary \\ \hline
297 & The Greek Philosophers, Vol. 2 & Alfred William Benn & Non-Literary \\ \hline
298 & An Ethical Philosophy of Life Presented in Its Main Outlines & Felix Adler & Non-Literary \\ \hline
299 & A Beginner's History of Philosophy, Vol. 1 & Herbert Ernest Cushman & Non-Literary \\ \hline
300 & Towards the Great Peace & Ralph Adams Cram & Non-Literary \\ \hline
301 & Society: Its Origin and Development & Henry K. Rowe & Non-Literary \\ \hline
302 & Criminal Man, According to the Classification of Cesare Lombroso & Gina Lombroso & Non-Literary \\ \hline
303 & The Challenge of the Country: A Study of Country Life Opportunity & George Walter Fiske & Non-Literary \\ \hline
304 & Criminal Sociology & Enrico Ferri & Non-Literary \\ \hline
305 & Community Civics and Rural Life & Arthur William Dunn & Non-Literary \\ \hline
306 & Sociology and Modern Social Problems & Charles A. Ellwood & Non-Literary \\ \hline
307 & The Theory of the Leisure Class & Thorstein Veblen & Non-Literary \\ \hline

\end{longtable}

\fontsize{11}{11}\selectfont

\begin{longtable}{c|ll|ll}
    \caption{Means ($\pm$ SD) of the different text properties analyzed in the present study. Abbreviation: MTLD, Measure for textual lexical diversity. Asterisks indicate that results are different from literary and canonical texts, respectively, at *, $p<0.05$; **, $p<0.01$; and ***, $p<0.001$ (paired t-tests).}
\\
\hline
\textbf{} & \textbf{Literary} & \textbf{Non-Literary} & \textbf{Canonical} & \textbf{Non-Canonical} \\
 \hline
\endfirsthead

        Noun & 3.80 (1.21) & 5.76 (1.21)$^{***}$ & 4.11 (1.24) & 3.55 (1.11)
 \Tstrut\\
        Verb & 3.30 (0.90) & 3.22 (0.96) & 3.54 (0.83) & 3.10 (0.90) \\
        Adjective & 1.28 (0.48) & 2.06 (0.64)$^{***}$ & 1.45 (0.45) & 1.13 (0.45)$^{*}$
 \\
        Pronoun & 2.04 (0.57) & 1.01 (0.49)$^{***}$ & 2.25 (0.57) & 1.86 (0.50)$^{*}$  \\
        Sentence-Length & 20.91 (6.15) & 25.13 (6.18)$^{***}$ & 22.82 (5.83) & 19.37 (5.96)$^{*}$ \\
        MTLD & 48.34 (7.44) & 45.43 (8.56)$^{**}$ & 47.81 (6.08) & 48.76 (8.36) \\
        Topic Distribution & 0.70 (0.02) & 0.68 (0.03)$^{***}$ & 0.70 (0.02) & 0.71 (0.03) \\\hline
\end{longtable}